\documentclass[10pt,twocolumn,letterpaper]{article}

\usepackage[pagenumbers]{cvpr} 

\definecolor{cvprblue}{rgb}{0.21,0.49,0.74}
\usepackage[pagebackref=true,colorlinks,urlcolor=citypink,citecolor=cityblue,linkcolor=citypink]{hyperref}

\usepackage{graphicx}
\usepackage{subcaption}
\usepackage{tikz}

\usepackage{amsmath,amsfonts,bm}

\def\eqref#1{equation~\ref{#1}}

\def\1{\bm{1}}

\DeclareMathAlphabet{\mathsfit}{\encodingdefault}{\sfdefault}{m}{sl}
\SetMathAlphabet{\mathsfit}{bold}{\encodingdefault}{\sfdefault}{bx}{n}

\usepackage[dvipsnames]{xcolor}
\usepackage{wrapfig}
\usepackage{float,wrapfig}
\usepackage{subcaption}
\usepackage{algpseudocode} 
\usepackage[ruled,vlined]{algorithm2e} 

\definecolor{citypink}{RGB}{227, 108, 194}
\definecolor{cityblue}{RGB}{128, 159, 225}

\usepackage{indentfirst}
\usepackage{titletoc}
\usepackage{cuted}
\usepackage{capt-of}
\usepackage{textcomp, gensymb}
\usepackage{courier}
\usepackage{multirow}

\usepackage{url}
\usepackage{balance}

\newcommand{\OM}{MVPaint}

\def\paperID{*****} 
\def\confName{CVPR}
\def\confYear{2025}

\begin{document}
\title{
    \raisebox{-0.01\linewidth}{\includegraphics[width=0.8cm]{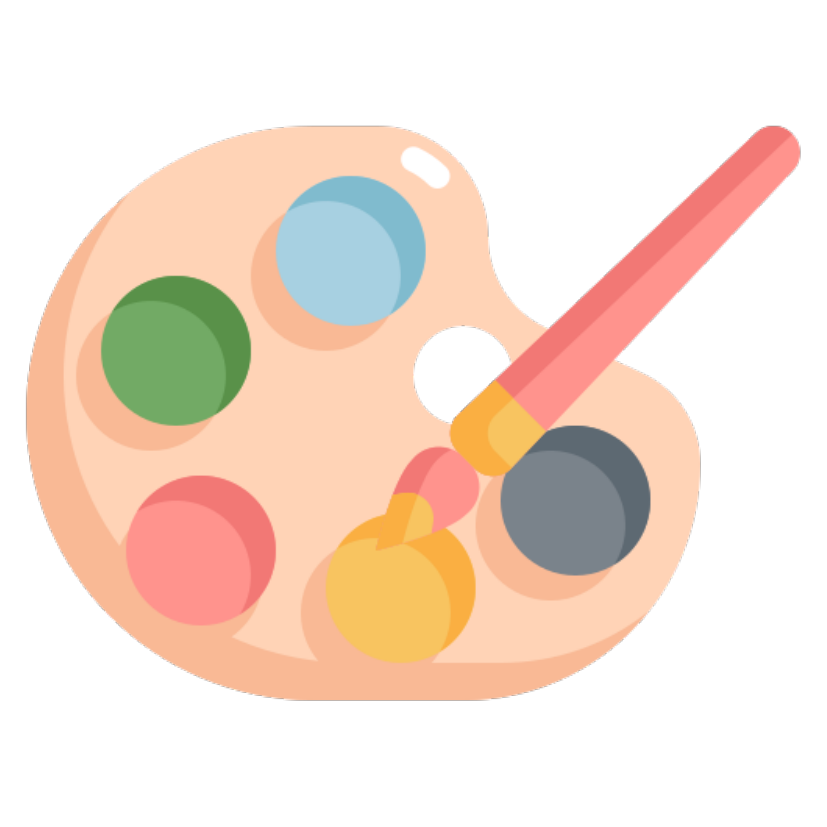}} 
    \textsf{\textcolor{citypink}{MVP}\textcolor{cityblue}{aint}}: 
    Synchronized Multi-View Diffusion for Painting Anything 3D
}
\author{
    Wei Cheng$^{1*\dagger}$
    \quad
    Juncheng Mu$^{2,4*}$
    \quad
    Xianfang Zeng$^{1}$
    \quad
    Xin Chen$^{1}$
    \quad
    Anqi Pang$^{1}$\\ \vspace{1mm}
    Chi Zhang$^{1}$ 
    \quad
    Zhibin Wang$^{1}$ 
    \quad
    Bin Fu$^{1}$ 
    \quad
    Gang Yu$^{1}$
    \quad
    Ziwei Liu$^{3}$
    \quad
    Liang Pan$^{2\ddagger}$
    \\ \vspace{1mm}
    $^{1}$ Tencent PCG
    \quad
    $^{2}$ Shanghai AI Laboratory
    \quad
    $^{3}$ S-Lab, NTU
    \quad
    $^{4}$ Tsinghua University \\ \vspace{-8mm}
    {\href{http://mvpaint.github.io}{\textbf{\texttt{http://mvpaint.github.io}}}}
}

\twocolumn[{
  \renewcommand\twocolumn[1][]{#1}
  \maketitle
  \begin{center}
  \vspace{5mm}
  \includegraphics[width=\textwidth]{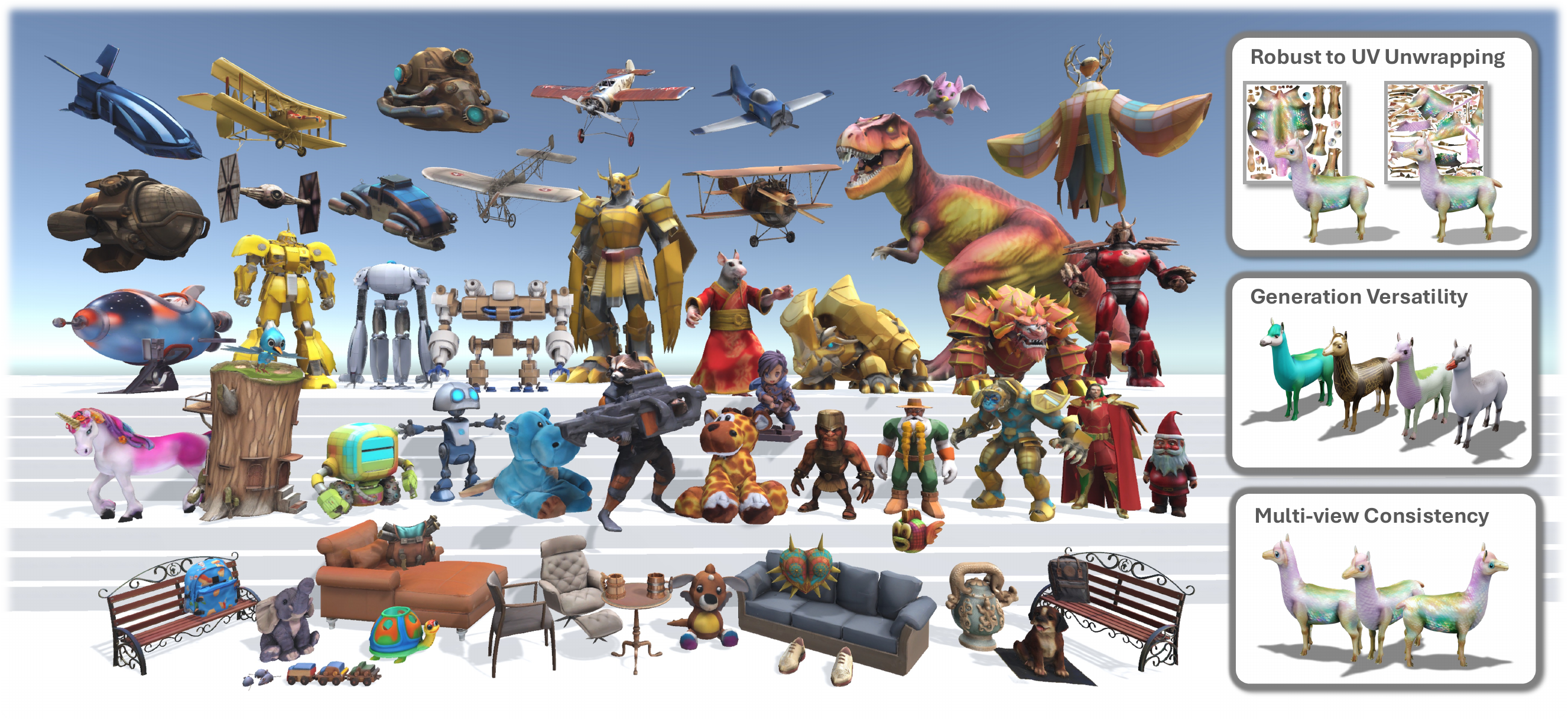}
  \captionof{figure}{\textsf{\textcolor{citypink}{MVP}\textcolor{cityblue}{aint}} generates \textbf{multi-view consistent} textures with \textbf{arbitrary UV unwrapping} and high \textbf{generation versatility}. }
  \label{fig:teaser}
  \end{center}
}]
{\let\thefootnote\relax\footnotetext{\noindent *~Equal Contribution, \dag~Project Lead, \ddag~Corresponding Author}}


\maketitle

\begin{abstract}

Texturing is a crucial step in the 3D asset production workflow, which enhances the visual appeal and diversity of 3D assets.
Despite recent advancements in Text-to-Texture (T2T) generation, existing methods often yield subpar results, primarily due to local discontinuities, inconsistencies across multiple views, and their heavy dependence on UV unwrapping outcomes.
To tackle these challenges, we propose a novel generation-refinement 3D texturing framework called \textsf{{\textcolor{citypink}{MVP}\textcolor{cityblue}{aint}}}, which can generate high-resolution, seamless textures while emphasizing multi-view consistency.
\OM{} mainly consists of three key modules.
\textbf{1) Synchronized Multi-view Generation (SMG)}. 
Given a 3D mesh model, \OM{} 
first simultaneously generates multi-view images by employing 
an SMG model,
which leads to coarse texturing results with 
unpainted parts due to missing observations.
\textbf{2) Spatial-aware 3D Inpainting (S3I)}.
To ensure complete 3D texturing, we introduce the 
S3I method,
specifically designed to effectively texture previously unobserved areas.
\textbf{3) UV Refinement (UVR)}.
Furthermore, \OM{} employs 
a UVR module to improve the texture quality in the UV space, 
which first performs a UV-space Super-Resolution, followed by a Spatial-aware Seam-Smoothing algorithm for revising spatial texturing discontinuities caused by UV unwrapping.
Moreover, we establish two T2T evaluation benchmarks: the Objaverse T2T benchmark and the GSO T2T benchmark, based on selected high-quality 3D meshes from the Objaverse dataset and the entire GSO dataset, respectively.
Extensive experimental results demonstrate that \OM{} surpasses existing state-of-the-art methods.
Notably, \OM{} could generate high-fidelity textures with minimal Janus issues and highly enhanced cross-view consistency.

\end{abstract}
\section{Introduction}

3D texture generation remains a complex and critical aspect of asset creation, especially valuable in applications like gaming, animation, and virtual/augmented/mixed reality.
Despite the scarcity of specialized 3D training data and the high computational demands of texture modeling, recent breakthroughs in text-to-image technologies~\cite{cascaded-diffusion,make-a-scene,Mastering-text-to-image-generation,LDM} have significantly advanced the field.
These technologies facilitate Text-to-Texture (T2T) generation~\cite{text2tex, Texture, paint3d, syncmvd} and their integration with 3D shapes~\cite{dreamfusion, dreamgaussian, wonder3d}, enhancing the visual diversity and realism of 3D models.
However, achieving consistent and seamless textures across various viewing angles remains challenging, often hampered by local discontinuities and cross-view inconsistencies.

Recently, many texture generation methods have focused on leveraging 2D diffusion priors for guiding the generation process, which often utilizes conditional controls~\cite{controlnet} (\eg, depth) to produce more fitting textures.
TEXTure~\cite{Texture} and Text2Tex~\cite{text2tex} sample a series of camera viewpoints for iteratively rendering depth maps, which are then used to generate high-quality images through a pre-trained depth-to-image diffusion model.
To avoid inconsistent textures due to using multiple independent generation processes, 
SyncMVD~\cite{syncmvd} introduced a method combining multi-view single-step denoising with UV space synchronization.
However, its reused attention process is limited to nearby views, which frequently leads to Janus problems.
Paint3D~\cite{paint3d} developed a coarse-to-fine texture generation strategy, starting with a coarse texture obtained by iteratively painting from camera viewpoints, followed by inpainting and super-resolution in UV space.
Similarly, Meta 3D TextureGen~\cite{meta3dtexgen} 
also employs UV position maps for UV inpainting and enhancement.
Despite achieving remarkable 3D texturing results,
both tools~\cite{paint3d, meta3dtexgen} depend heavily on continuous mesh UV unwrapping. 
Discontinuities in texture often arise in scenarios where the UV atlases are randomly packed within the UV images.
Consequently, 
many challenges remain in 3D texturing:
\textbf{Multi-View Consistency}: Ensuring consistency across multiple viewpoints to prevent local style discontinuities and the presence of numerous seams.
\textbf{Diverse Texture Details}: Avoiding overly smooth textures that lack detail, while aiming for high-resolution outputs.
\textbf{UV Unwrapping Robustness}: Developing a method that does not rely heavily on UV unwrapping results to achieve robust automated generation.

To address these challenges, we propose \textsf{\textcolor{citypink}{MVP}\textcolor{cityblue}{aint}}, a coarse-to-fine 3D texture generation framework capable of producing high-fidelity, seamless 3D textures while ensuring multi-view consistency and reducing dependence on UV unwrapping quality.
\OM{} mainly consists of three stages for texture generation.
(1) First, we employ the \textbf{Synchronized Multi-view Generation} (SMG) model that uses a multi-view diffusion model with cross-attention~\cite{attention} and UV synchronization to initiate 3D texture generation conditioned on a given textural instruction, which effectively avoids the Janus problem and produces highly consistent multi-view images at low resolution.
Following, we upsample and refine the coarse multi-view images by adding vivid texture details, subsequently projecting them into UV space (1K resolution) for further enhancement.
(2) Second, we propose the \textbf{Spatial-aware 3D Inpainting} (S3I) method to ensure complete 3D texturing, particularly for areas that were not observed in the first stage.
Specifically, S3I resolves the inpainting process in 3D space by considering the spatial relations among 3D points uniformly sampled from mesh surfaces.
(3) Third, we introduce a \textbf{UV Refinement} (UVR) module, comprising a series of tailored texture enhancement operations in UV space.
UVR first employs a super-resolution module to upscale the UV map to 2K resolution.
Afterward, we introduce a Spatial-aware Seam-smoothing Algorithm (SSA) to revise spatial discontinuous textures, especially for repairing the seams caused by UV unwrapping.
Consequently, high-quality 3D UV textures could be obtained. 

To facilitate the evaluation of T2T generation, we establish two benchmarks: the Objaverse~\cite{objaverse} T2T benchmark and the GSO~\cite{gso} T2T benchmark.
The Objaverse T2T benchmark comprises 1000 high-quality 3D meshes curated from the Objaverse dataset. 
Given that most T2T models are trained on a subset of the Objaverse dataset, we further establish the GSO T2T benchmark, which leverages all 1032 3D models from the GSO dataset to assess the generalizability of T2T models.
For each 3D mesh, textual annotations are generated utilizing a large language model (LLM). 
Extensive experimental results on the Objaverse~\cite{objaverse} and the GSO~\cite{gso} T2T benchmarks demonstrate that \OM{} could outperform existing State-of-The-Art (SoTA) methods for 3D texture generation.
We would like to emphasize that \OM{} is a robust 3D texturing method, significantly reducing occurrences of failed generations, such as missing areas, large inconsistencies, over-smoothness, and Janus issues.
Qualitative texturing results of \OM{} could be visualized in Fig.~\ref{fig:teaser}.

Our contributions could be summarized as follows:
\textbf{1)} We propose a robust 3D texturing framework, entitled \OM{}, for generating diverse, high-quality, seamless 3D textures while ensuring multi-view consistency.
\textbf{2)} Various 3D texturing models, operations, and strategies, including SMG, S3I, and UVR modules, have been proposed, studied, and utilized in this work. We believe these contributions will significantly advance future research in 3D texture generation.
\textbf{3)} We conduct extensive experiments on the Objaverse and the GSO T2T benchmark, demonstrating that \OM{} achieves impressive 3D texture generation results, surpassing existing SoTA methods.

\begin{figure*}
    \centering
    \includegraphics[width=\linewidth]{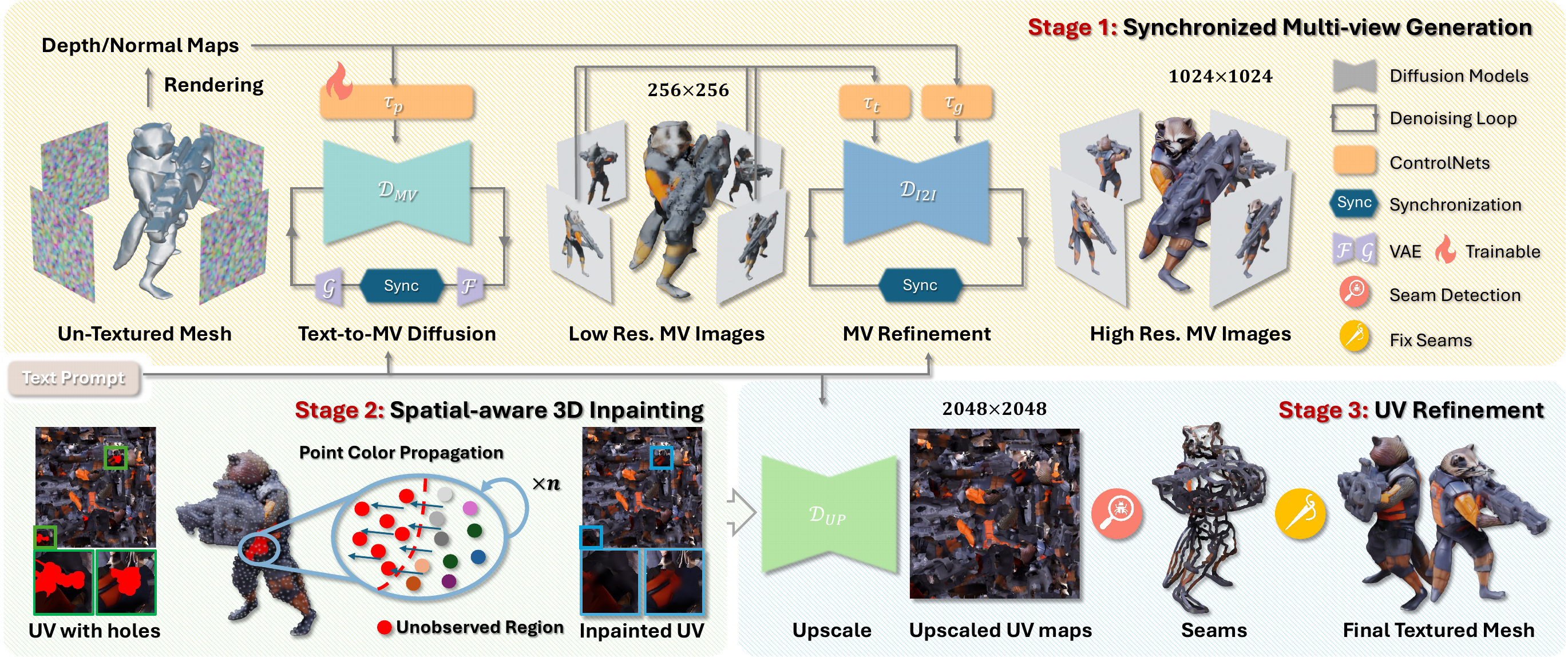}
    \vspace{-5mm}
    \caption{
    \textbf{The Framework Overview of \OM{}.}
    Given an input mesh, Stage 1 of \OM{} utilizes a synchronized multi-view generation (\textbf{SMG}) model,
    consisting of a control-based T2MV model and an I2I model,
    for 3D texture initialization.
    In Stage 2, the synchronized views are reprojected back to UV space, where inpainting is performed on the 3D point cloud to fill the holes (shown in red dots), hence completing the UV map. 
    In Stage 3, the completed UV map undergoes super-resolution, adding finer details, followed by seam detection and 3D-aware smoothing to achieve a complete, seamless, and multi-view consistent 3D texture.
    }
    \label{fig:pipeline}
\end{figure*}

\section{Related Work}

\noindent
\textbf{Multi-View Generation.} 
The generation of coherent multi-view images from diverse inputs, such as text, images, and meshes, has become a crucial research area. This enables the creation of 3D assets with consistent appearances across various perspectives. 
Groundbreaking studies like Zero-1-to-3~\cite{zero123} and Consistent-1-to-3~\cite{consistent123} have employed viewpoint-conditioned diffusion models to synthesize novel views of objects, ensuring seamless transitions between the generated perspectives. 
Building upon these advancements, Zero123++~\cite{zero123++} and MVDream~\cite{mvdream} have adopted an approach that tiles multiple view images into a single canvas for generation, respectively producing consistent hexa-view and tetra-view images.

\noindent
\textbf{3D Texture Generation.}
Traditionally, texture generation relies on manual or procedural techniques~\cite{old2,old3,old4,old5}, which were effective for basic applications but lacked complexity.
The introduction of global optimization techniques~\cite{old6,old1} allows for more detailed textures that better matched 3D model geometries.
AI-based 3D texture generation is initially dominated by generative adversarial networks (GANs)~\cite{GAN,CGAN,DCGAN,StyleGAN}, and then the focus has shifted towards latent diffusion models (LDM)~\cite{DDPM,LDM}, with models like Stable Diffusion~\cite{LDM, sdxl} showing promising results.  
With large-scale text-image prior models,
texture generation methods~\cite{Texture,syncmvd,it3d} leverage text-image correlations to update rendered 3D views by extracting gradients from CLIP~\cite{clip-forge,clip-mesh,CLIP}.
Iterative methods~\cite{Texture,flashtex,easi} enhance texture quality and consistency across 3D models by rendering depth or normal maps, with the help of ControlNet~\cite{controlnet}.
Other methods~\cite{paint3d,meta3dtexgen} generate multi-view images with depth or normal conditioned sparse views and then apply inpainting and refinement directly on a UV map with a position-map-controlled diffusion network. 
However, it is challenging for UV diffusion models to directly generate correct and 3D-continuous texture patches, as they are frequently packed to separate UV regions.
Recently, multi-path diffusion~\cite{multidiffusion} synchronizes the wrapped latent~\cite{syncmvd} or images~\cite{texpainter,flexitex} during multiple single-view DDIM~\cite{ddim} processes. 
Nonetheless, these methods~\cite{syncmvd,flexitex} tend to get trapped to the multi-face issues also known as the Janus problem which is a typical phenomenon in 3D generation using 2D priors. 
More discussions are provided in Sec.~\ref{sec:detailed_related}.

\section{Our Approach}

Given an untextured mesh $\mathbf{M} = (\mathrm{V}, \mathrm{F})$ and a texture prompt $c$, where $\mathrm{V} = \{v_i\}$ representing the set of 3D vertex $v_i \in \mathbb{R}^3$ and $\mathrm{F} = \{f_i\}$ representing the set of triangular faces each defined by a triplet of vertices, \OM{} aims to generate a high-quality texture map (2K level) represented as a multi-channel UV image $\mathbf{T} \in \mathbb{R}^{H\times W\times C}$ conditioned on the texture prompt $c$.
To achieve multi-view consistency, high-quality, and seamless 3D textures, 
\OM{} utilizes three major stages, including 
\textbf{1)}
Synchronized Multi-view Generation (SMG) model - for simultaneously generating dense view images as the initial texturing (see Sec.~\ref{Stage One}); 
\textbf{2)}
Spatial-aware 3D Inpainting (S3I) model - to inpaint and enhance the texturing based on spatial relations (see Sec.~\ref{Stage Two});
and 
\textbf{3)}
UV Refinement (UVR) module - to conduct upsampling and refinement for generating the final UV texture (see Sec.~\ref{Stage Three});.
The overview of \OM{} is illustrated in Fig.~\ref{fig:pipeline}.

\subsection{Synchronized Multi-View Generation}
\label{Stage One}

Building on the success of using 2D diffusion priors in recent Text-to-Image (T2I) models~\cite{cascaded-diffusion,make-a-scene,Mastering-text-to-image-generation,LDM}, many 3D texturing methods~\cite{Texture,text2tex,syncmvd,paint3d} initialize their textures using depth-conditioned 2D generation models.
Specifically, they typically begin by rendering depth maps from various viewpoints sampled around the given 3D mesh model $\mathbf{M}$, then using a pre-trained depth-to-image diffusion model to generate multi-view images based on the text instructions.
Despite leveraging 2D generation priors, existing methods often overlook 3D priors during generation, leading to low-quality multi-view results plagued by issues such as multi-view inconsistencies, the Janus problem, and over-smoothed textures lacking vivid details.

\begin{figure}[t]
    \centering
    \vspace{-3mm}
    \includegraphics[width=\linewidth]{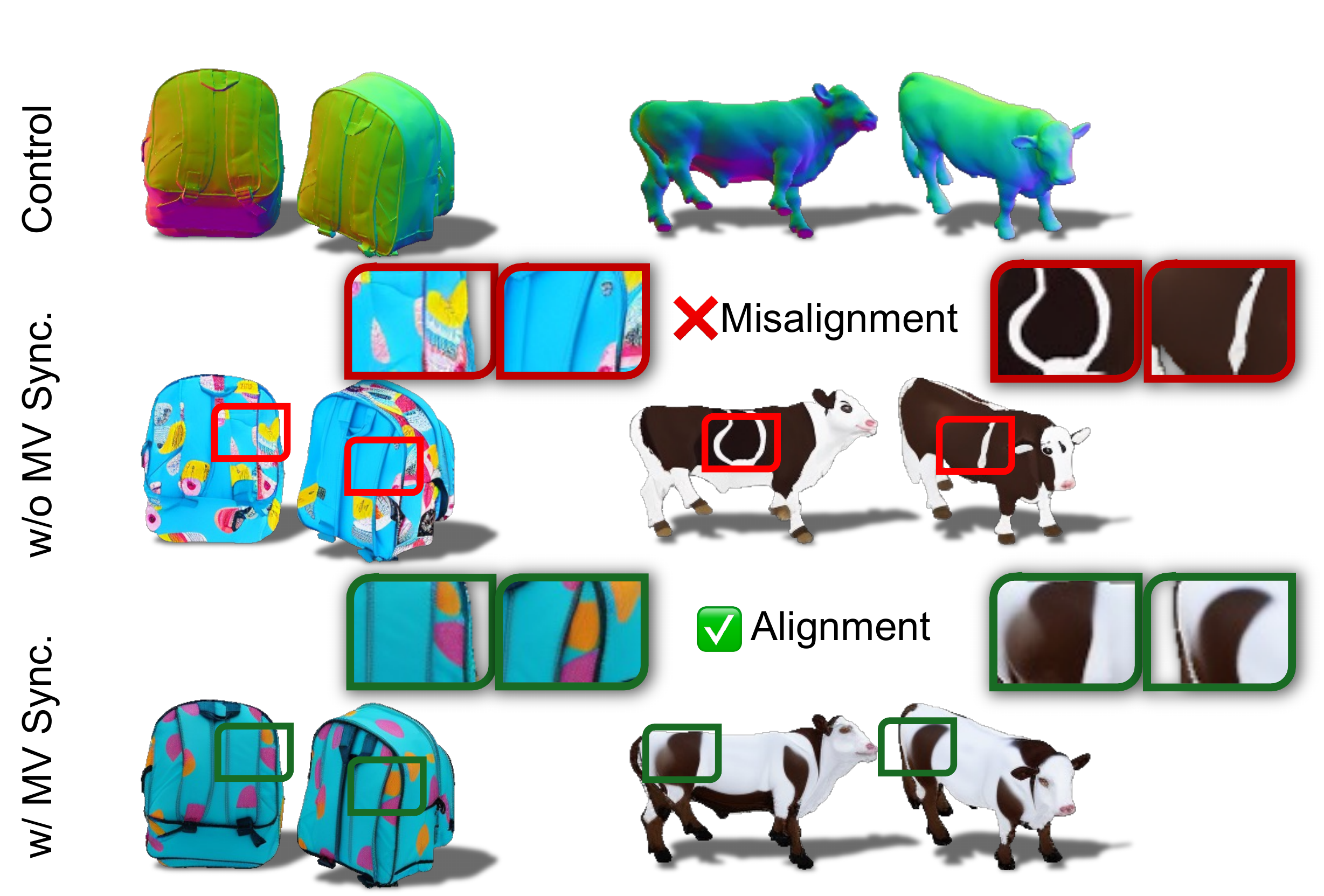}
    \vspace{-7mm}
    \caption{
    \textbf{The Effectiveness of Synchronization on Multi-view Image Generation.}
    Although T2MV models generate Janus-problem-free results, they still suffer from texture misalignment from different views. 
    In contrast, the proposed SMG model can effectively enforce multi-view consistency for T2MV generation.
    } 
    \label{fig:stage1_sync}
\end{figure}

\noindent
\textbf{Text-to-Multi-View for 3D Texture Generation.}
In contrast, we utilize a Text-to-Multi-View (T2MV) diffusion model $\mathcal{D}_{\text{MV}}$, where the multi-view prior acts as a generalizable 3D prior~\cite{mvdream}, to generate consistent multi-view images in a single forward process conditioned on the text instruction $c$.
In particular, we leverage the T2MV model $\mathcal{D}_{\text{MV}}$ by training a control model~\cite{controlnet} $\tau_p$ 
to guide the generation process using the depth or normal map $\mathbf{P}_{\{1,\,2,\,\cdots,\,N\}}^l$ (abbr. as $\mathbf{P}_\mathrm{N}^l$) from the corresponding view.
Formally, the generation of initial low-resolution multi-view images $\mathbf{I}_{\{1,\,2,\,\cdots,\,N\}}^l$ (abbr. as $\mathbf{I}_\mathrm{N}^l$) could be formulated as:
\begin{equation}
    \mathbf{I}_\mathrm{N}^l = \mathbf{I}_{\{1,\,2,\,\cdots,\,N\}}^l = \mathcal{D}_{\text{MV}}(z^{\text{MV}}, \,c, \,\mathbf{P}_\mathrm{N}^l; \,\tau_p),
\end{equation}
where $z^{\text{MV}}$ is latent for MV generation.

\noindent
\textbf{Multi-View Synchronization.}
Existing T2I-model-based texturing methods~\cite{multidiffusion,syncdiffusion,syncmvd} enhance the alignment of texture predictions from different views by integrating their denoising processes on the latent according to the shared UV space.
Although significant differences across different views could lead to overly smoothed textures, the synchronization operation could eliminate minor discrepancies by adjusting the generation results of different views based on the same UV map.
Unfortunately, the latent of a T2MV model $\mathcal{D}_{\text{MV}}$ usually has a low resolution (\emph{e.g.}, $32\times 32$), which complicates establishing a robust mapping to the UV space, especially when UV unwrapping is complex, hence hindering the effective capture of intricate visual relationships across different views.

Rather than synchronizing in latent space, we propose the 
Synchronized Multi-view Generation (\textbf{SMG}) model, 
which fuses the multi-view diffusion path by aligning the multi-view generation in the decoded image domain.
Given the latent $z^{\text{MV}}_t$ of $\mathcal{D}_{\text{MV}}$ at time step $t$, 
a clean intermediate state $z^{\text{MV}}_{0|t}$ (denoted as $z^{\prime\,\text{MV}}$ in the following for simplicity) is obtained by removing the noise from $z^{\text{MV}}_t$.
Afterward, $z^{\prime\,\text{MV}}$ is 
converted into image space 
$\mathbf{I}^{\prime}_\mathrm{N}$
with a larger resolution (\emph{e.g.}, $256\times 256$) 
by using the pretrained decoder $\mathcal{G}_{\psi}(\cdot)$ of the VAE in $\mathcal{D}_{\text{MV}}$.
Subsequently, we generate the synchronized UV map $\textbf{T}^{\prime}_{\text{sync}}$ by fusing the multi-view generation with inverse UV mapping, applying weights based on the cosine angle between the view direction $\mathbf{v}_i$ and the surface normal $\mathbf{n}_{uv}$ represented in the UV space.
Afterward, the synchronized multi-view images 
$\mathbf{I}^{\prime}_{\mathrm{N}_\text{sync}}$
are generated by rasterizing 
the UV map, which is then encoded as synchronized latent $z^{\prime\,\text{MV}}_{\text{sync}}$ using the pretrained encoder $\mathcal{F}_{\phi}(\cdot)$ of the VAE in $\mathcal{D}_{\text{MV}}$.
Consequently, the multi-view consistency could be enhanced, and the synchronized multi-view images $\mathbf{I}^{l}_{\mathrm{N}_\text{sync}}$ could be obtained as follows:
\begin{align}
    &\;
    \mathbf{I}^{l}_{\mathrm{N}_\text{sync}} = \mathcal{D}_{\text{MV}}(\{z^{\text{MV}}; z^{\prime\,\text{MV}}_{\text{sync}}\}, \,c, \,\mathbf{P}_\mathrm{N}^l; \,\tau_p) \,,
    \;\;\;\;\;\;\;\;
    \\
    \text{where } 
    &\left\{
    \begin{aligned}
        &z^{\prime\, \text{MV}}_{\text{sync}} = \mathcal{F}_{\phi}(\mathbf{I}^{\prime}_{\mathrm{N}_\text{sync}}) ,\;\;
        \mathbf{I}^{\prime}_{\mathrm{N}_\text{sync}} = \mathcal{R}(\textbf{T}^{\prime}_{\text{sync}})\, , \\
        &\mathbf{T}^{\prime}_{\text{sync}} = \sum_i^N{\cos(\mathbf{v}_i, \mathbf{n}_{uv})}\mathbf{T}^{\prime}_i ,
        \\
        &\mathbf{T}^{\prime}_i = \mathcal{R}^{-1}(\mathbf{I}^{\prime}_i), \;\; \mathbf{I}^{\prime}_{\mathrm{N}} = \mathcal{G}_{\psi}(z^{\prime\, \text{MV}})\, ,
        \label{equ:mapping}
    \end{aligned}
    \right.
\end{align}
$\mathcal{R}(\cdot)$ and $\mathcal{R}^{-1}(\cdot)$ denote the UV rasterization and UV mapping function, respectively. 
Notably, using a single synchronization step could be adequate, while many synchronized diffusion steps may result in unstable generations.
The effectiveness of synchronization could be visualized in Fig.~\ref{fig:stage1_sync}.

\noindent
\textbf{Synchronized Refinement.} 
To enhance and upscale multi-view images $\mathbf{I}^{l}_{\mathrm{N}_\text{sync}}$, 
we utilize an Image-to-Image (I2I) generation module $\mathcal{D}_\text{I2I}(\cdot)$ for synchronized texture refinement,
which generates high-fidelity images $\mathbf{I}^{h}_{\mathrm{N}_\text{sync}}$ at a higher resolution (1K level) while preserving multi-view consistency.
Specifically, we leverage two pretrained control models~\cite{LDM,controlnet}, including a texture refinement model $\tau_t$ and a geometry enforcement model $\tau_g$, which adopts the synchronized denoising mechanism~\cite{syncmvd} to ensure consistency across different views.
Formally, the high-quality multi-view image refinement could be formulated as:
\begin{equation}
    \mathbf{I}^{h}_{\mathrm{N}_\text{sync}} = \mathcal{D}_\text{I2I}(z^{\text{I}}, \mathbf{I}^{l}_{\mathrm{N}_\text{sync}}, \mathbf{P}_\mathrm{N}^h, s_t, s_g; \tau_t, \tau_g),
\end{equation}
where $z^{\text{I}}$ is a random initialized latent for single image generation, $\mathbf{P}_\mathrm{N}^h$ is the set of $N$ high-resolution geometric (\emph{e.g.}, depth or normal) maps, and $s_t, s_g$ is the user-defined strength of the two control models.

\begin{figure}[t]
    \centering
        \centering
        \includegraphics[width=0.95\linewidth]{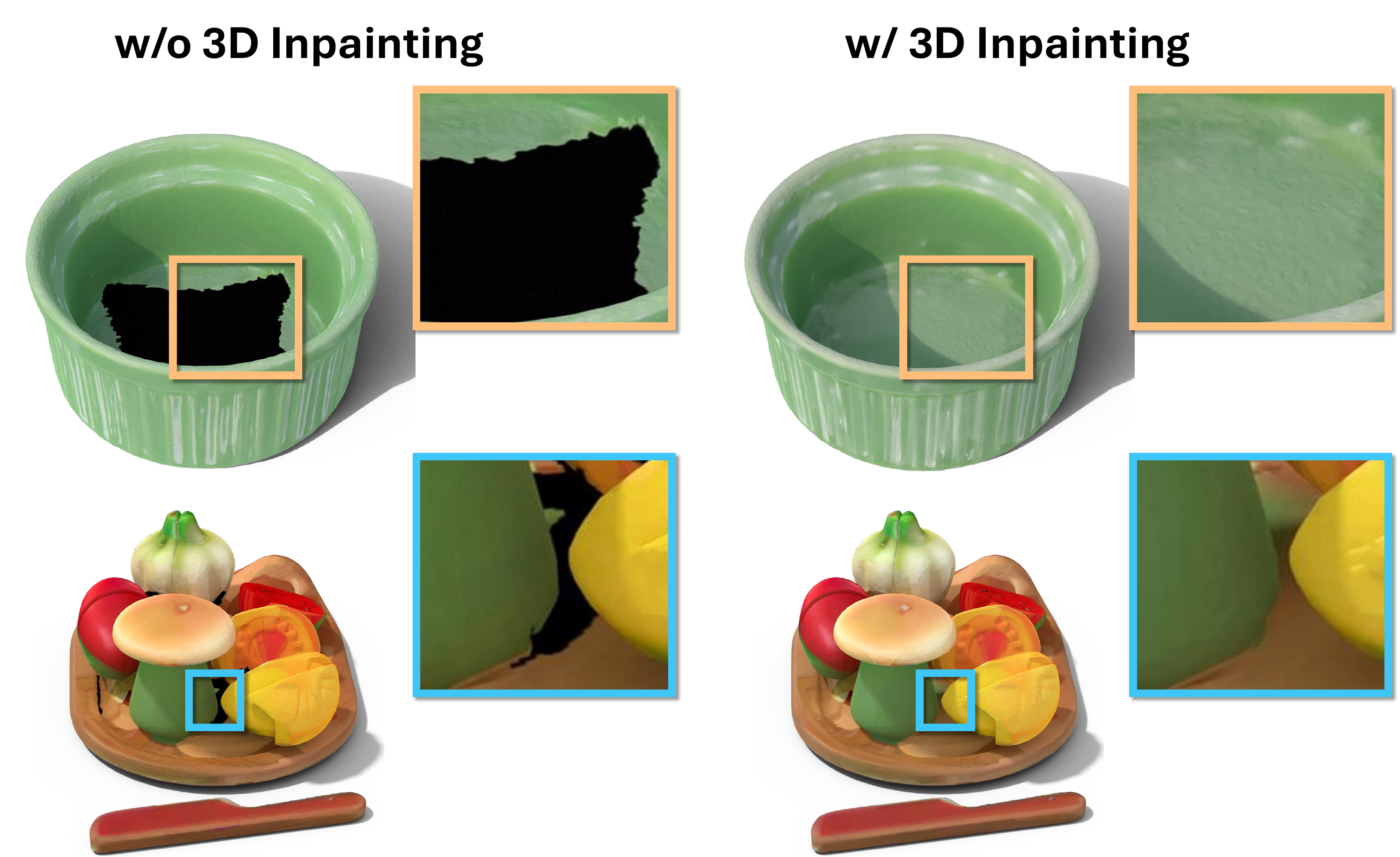}
        \vspace{-3mm}
        \caption{
        \textbf{Spatial-aware 3D Inpainting} 
        could effectively accomplish texture completion for 3D structures with complex geometries and large unobserved areas.
        }
        \label{fig:ablation_inpainting}
\end{figure}

\subsection{Texture Inpainting in 3D Space}
\label{Stage Two}
Although high-quality multi-view images $\mathbf{I}^{h}_{\mathrm{N}_\text{sync}}$ cover most of the mesh surface, there are still unobserved regions that need to be inpainted.
After performing UV mapping $\mathcal{R}^{-1}(\cdot)$, multi-view images $\mathbf{I}^{h}_{\mathrm{N}_\text{sync}}$ are projected into the UV space, achieving an incomplete UV map $\mathbf{T}_i$.
To address the artifacts caused by complex self-occlusion issues during UV mapping $\mathcal{R}^{-1}(\cdot)$, we refine the projection area by identifying and limiting the regions affected by occlusion (check Sec.~\ref{subsec: URA} for details).
For UV map completion, existing methods~\cite{paint3d, syncmvd} mostly perform inpainting directly in UV space.
However, adjacent 3D areas are frequently mapped to non-adjacent 2D regions within $\mathbf{T}_i$, which becomes more pronounced when $\mathbf{T}_i$ is highly fragmented.

\noindent
\textbf{3D Point Cloud Inpainting.}
In light of this, we propose a Spatial-aware 3D Inpainting (\textbf{S3I}) module (shown in Stage 2 of Fig.\ref{fig:pipeline}) to inpaint texture in the 3D space, 
which generates a full-coverage 3D texture $\mathbf{T}_c$ conditioned on the incomplete texture $\mathbf{T}_i$,
while enforcing 3D geometry-aware spatial consistency.
Specifically, we first generate a dense colored point cloud $\mathbf{P}_{uv} \in \mathbb{R}^{N_{uv} \times 6}$ by concatenating the 3D point coordinates generated based on each pixel from valid regions in $\mathbf{T}_i$ and their corresponding RGB values.
Note that unpainted pixels from valid UV areas will also be used for generating 3D points $\mathbf{P}_{u} \subset \mathbf{P}_{uv}$ with zero-initialized color vector (\emph{i.e.}, $(0, 0, 0)$).
Accordingly, inpainting the UV texture could be reformulated as predicting a suitable color vector for each point $\mathbf{p} \in \mathbf{P}_{u}$ conditioned on the set of colored 3D points $\mathbf{P}_{v} \subset \mathbf{P}_{uv}$ generated from the visible area in $\mathbf{T}_c$. 
We highlight that the learning-free approach S3I is unaffected by the UV unwrapping results.

\noindent
\textbf{Spatial-aware Color Propagation.}
To address the 3D point inpainting problem, we propose a Spatial-aware Color Propagation (\textbf{SCP}) algorithm, 
which iteratively propagates the color value from $\mathbf{P}_{v}$ to $\mathbf{P}_{u}$.
In each iteration, the $k$-nearest neighbors $\mathbf{N}^{\mathbf{p}_i}_{\mathrm{k}} = \{ \mathbf{q}_j\in \mathbf{P}_{v}\, | \, 1\le j \le k \}$ of each point $\mathbf{p}_i \in \mathbf{P}_{u}$ are selected, and the color vector of $\mathbf{p}_i$ is then estimated by applying a weighted sum of the color vectors from each neighbor within $\mathbf{N}^{\mathbf{p}_i}_{\mathrm{k}}$.
For each neighboring point $\mathbf{q}_j \in \mathbf{N}^{\mathbf{p}_i}_{\mathrm{k}}$, the aggregation weight $w_j$ is computed by considering both the Euclidean distance $d_j$ and the surface normal similarity between $\mathbf{q}_j$ and $\mathbf{p}_i$:
\begin{equation}
    w_j = \frac{1/d_j}{\sum_j {1/d_j}} \times f(\mathbf{n}_j \cdot \mathbf{n}_i),
\end{equation}
where $\mathbf{n}_j$ and $\mathbf{n}_i$ are surface normal of $\mathbf{q}_j$ and $\mathbf{p}_i$, respectively.
$f(\cdot)$ is a robust mapping function, defined as:
\begin{equation}
    f(x) = 
\begin{cases} 
1 \times 10^{-8}, & \text{if }\; -1 \leq x < 0.5, \\
x, & \text{if }\; 0.5 \leq x < 0.9, \\
10, & \text{if }\; 0.9 \leq x \leq 1.
\end{cases}
\end{equation}
After $n$ iterations, all 3D points will be painted, resulting in the full-coverage UV texture $\mathbf{T}_c$ (1K level).
Please refer to the supplementary (Sec.~\ref{subsec: S3I}) for algorithm details.
The effectiveness of S3I could be visualized in Fig.~\ref{fig:ablation_inpainting}.

\begin{figure}[t]
        \centering
        \includegraphics[width=1\linewidth]{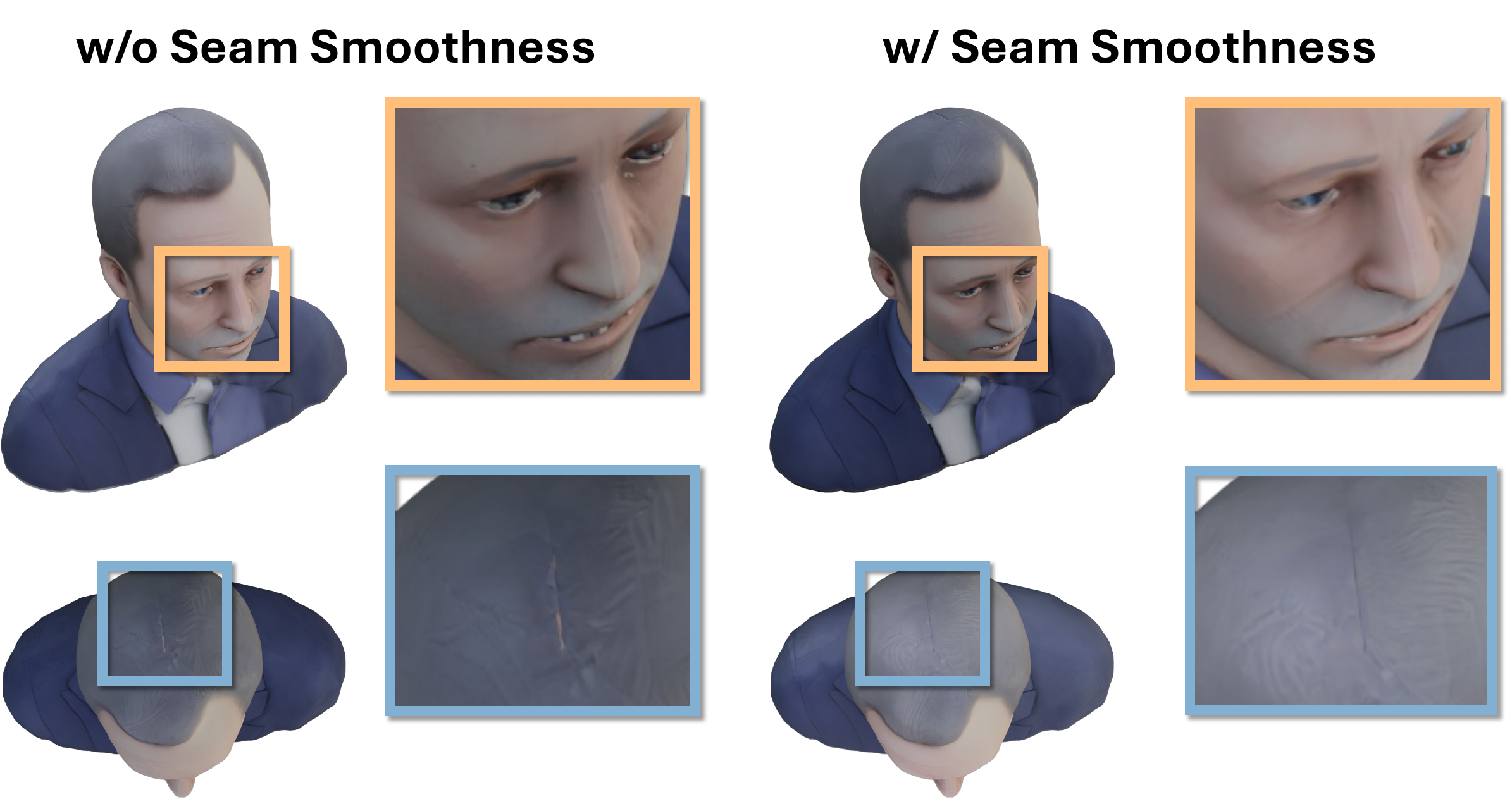}
        \vspace{-5mm}
        \caption{
        \textbf{Spatial-aware Seam-smoothing Algorithm} 
        could revise texture seams from 2D UV unwrapping by smoothing color vectors using their 3D neighbors.
        }
        \label{fig:ablation_seam-smoothing}
\end{figure}

\begin{figure*}[t]
    \centering
    \includegraphics[width=\linewidth]{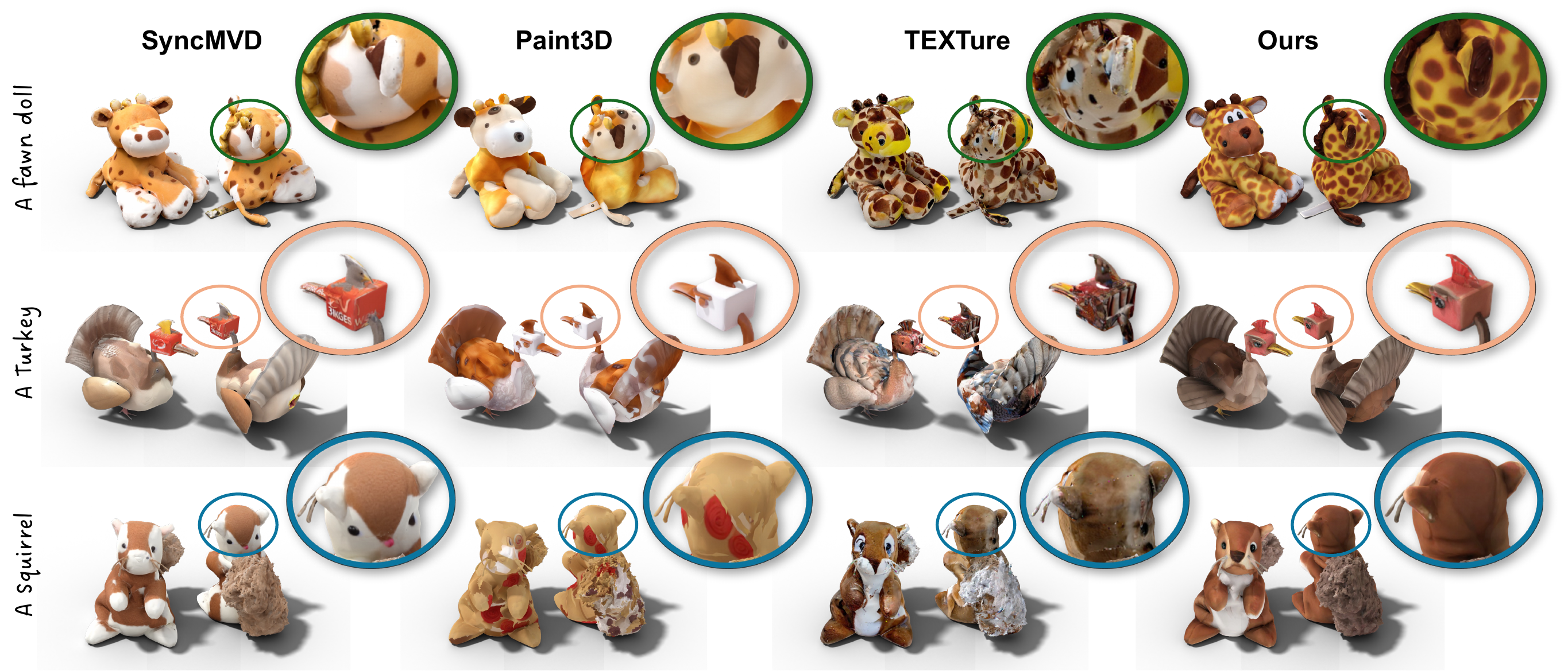}
    \vspace{-7mm}
    \caption{
    \textbf{Qualitative Results on Text-conditioned 3D Texture Generation.}
    \OM{} could constantly generate high-quality 3D textures, while existing methods frequently provide flawed results.
    Note that the input text prompts are simplified for better presentation.
    }
    \label{fig:qua}
\end{figure*}

\subsection{UV Refinement} \label{Stage Three}
Even though a full-coverage UV texture $\mathbf{T}_c$ has been acquired, texture details may appear flawed caused by projection errors during $\mathcal{R}^{-1}(\cdot)$ and the interpolation filling process.
To achieve a high-quality UV map $\mathbf{T}$,
we propose a UV Refinement (\textbf{UVR}) module (shown in Stage 3 of Fig.\ref{fig:pipeline}), mainly consisting of 
1) a super-resolution module for UV texture upsampling and refinement;
2) and a spatial-aware seam-smoothing algorithm for repairing the texture seams caused by UV upsampling.

\noindent
\textbf{UV Space Super-Resolution.} 
In order to generate more aesthetically pleasing and higher-resolution texture mappings, we perform super-resolution on $\mathbf{T}_c$ in UV space. 
We use an Image-to-Image Upscale (UP) diffusion model $\mathcal{D}_\text{UP}$ for super-resolution in UV space with an upscale model $\tau_\text{up}$: 
\begin{equation}
    \textbf{T}_\text{up} = \mathcal{D}_\text{UP}(z^{\text{UV}}, \textbf{T}_c; \tau_\text{up}),
\end{equation}
where $z^{\text{UV}}$ is a random initialized latent for high-resolution UV generation.
The upscale model $\tau_\text{up}$ can be selected as either a tiling control network or an upscale network. 
Please refer to Sec.~\ref{subsec:UV_tiling} for detailed model descriptions and distinctions.

\noindent
\textbf{Spatial-aware Seam-smoothing Algorithm.} 
Although more delicate and intricate textures could be received in the upsampled UV map $\mathbf{T}_{\text{up}}$, the discontinuities caused by UV unwrapping frequently lead to abrupt changes.
To resolve this problem, we introduce a Spatial-aware Seam-smoothing Algorithm (\textbf{SSA}) to revise the seams in 3D point cloud space.
First, we extract the binary image $\mathbf{T}_{\text{valid}}$ to mark the valid pixels from $\mathbf{T}_{\text{up}}$, based on which we perform connectivity analysis and edge extraction for detecting the seam mask $\mathbf{m}_{\text{seam}}$. 
Similar to the projection operations in S3I, we resample $\mathbf{T}_{\text{up}}$ into 3D colored point cloud $\mathbf{P}_{\text{up}} \in \mathbb{R}^{N_2 \times 6}$, followed by applying the SS algorithm for seam repairing.
Specifically, we extract subsets $\mathbf{P}_{\text{seam}}$ and $\mathbf{P}_{\text{n-seam}}$ from $\mathbf{P}_{\text{up}}$ using the seam mask $\mathbf{m}_{\text{seam}}$. 
Afterward, we construct a kd-tree with $\mathbf{P}_{\text{n-seam}}$ and then refine $\mathbf{P}_{\text{seam}}$ with neighboring points from $\mathbf{P}_{\text{n-seam}}$.
Finally, by calculating the normal vector cosine similarity and the distance for weighted coloring, we obtain the final high-quality seamless texture $\mathbf{T}$ (2K level). 
The effectiveness of SSA are illustrated in Fig.~\ref{fig:ablation_seam-smoothing}.
More algorithm details are provided in Sec.~\ref{subsec: SSA}.

\section{Experiments}
In this section, extensive experiments have been conducted to evaluate the effectiveness of \OM in generating high-quality 3D textures from texture instructions.

\noindent
\textbf{Implementation Details.}
We utilize the MVDream~\cite{mvdream} as the base model of T2MV $\mathcal{D}_{\text{MV}}$ of SMG, and we add the control module $\tau_{p}$ and train it with the same training scheme of ControlNet~\cite{controlnet}.
Different from other controlled MVDream methods~\cite{mvcontrol} which only controls a single view, we densely control multi-view for better shape alignment. 
For synchronized refinement, we choose the SDXL~\cite{sdxl} as the based model for I2I refinement $\mathcal{D}_{\text{I2I}}$,
where two pre-trained ControlNets~\cite{tile_sdxl,depth_sdxl}, $\tau_t$ and $\tau_g$ are deployed.
During refinement, the per-view latents $z^{\textbf{I}}$ are with $128\times128$ resolution, and they are synchronized on $\textbf{T}_{\text{sync}}$ with $512\times512$ resolution. 
In all SMG processes, models are worked on $N=8$ views with evenly distributed azimuth angle and interleaved elevation of $\pm 30$\textdegree.

\begin{table}
   \caption{
   \textbf{Quantitative Results on the Objaverse T2T Benchmark.}
   } 
   \vspace{-4mm}
   \label{tab:method_objaverse}
   \small
   \centering
   \resizebox{0.48\textwidth}{!}{
       \begin{tabular}{l|ccc|ccc}
       \toprule
        \multirow{2}{*}{\textbf{Method}} & \multirow{2}{*}{\textbf{FID$\downarrow$}} & \multirow{2}{*}{\textbf{KID}$\downarrow$} & \multirow{2}{*}{\textbf{CLIP$\uparrow$}} & \multicolumn{3}{c}{\textbf{User Study} } \\
        & & & & \textbf{Overall $\uparrow$} & \textbf{Seamless$\uparrow$} & \textbf{Consistency$\uparrow$} \\ 
       \midrule
        TEXTure~\cite{Texture}  & 28.03 & 7.60 & \textbf{20.30}  & 3.81 & 3.66 & 3.31 \\ 
        
        Paint3D~\cite{paint3d}  & 25.28 & 5.19 & 19.27  & 3.85 & 3.36 & 3.51  \\ 
        
        SyncMVD~\cite{syncmvd}  & 26.99 & 5.72 & 20.19  & 3.96 & 3.85 & 3.59 \\ 
        \midrule
        Ours     & \textbf{20.89} & \textbf{3.45} & 19.87  & \textbf{4.19} & \textbf{4.25} & \textbf{3.98}  \\ 
        \bottomrule
       \end{tabular}
   }
\end{table}
\begin{table}
    \vspace{-3mm}
   \caption{
   \textbf{Quantitative Results on the GSO T2T Benchmark.}
   } 
   \vspace{-4mm}
   \label{tab:method_gso}
   \small
   \centering
   \resizebox{0.48\textwidth}{!}{
       \begin{tabular}{l|ccc|ccc}
       \toprule
        \multirow{2}{*}{\textbf{Method}} & \multirow{2}{*}{\textbf{FID$\downarrow$}} & \multirow{2}{*}{\textbf{KID}$\downarrow$} & \multirow{2}{*}{\textbf{CLIP$\uparrow$}} & \multicolumn{3}{c}{\textbf{User Study} } \\
        & & & & \textbf{Overall $\uparrow$} & \textbf{Seamless$\uparrow$} & \textbf{Consistency$\uparrow$} \\ 
       \midrule
        TEXTure~\cite{Texture}  & 24.76  & 5.50  & \textbf{23.44} & 3.78 & 3.97 & 3.65 \\ 
        
        Paint3D~\cite{paint3d}  & 37.29  & 10.24  & 21.21  & 3.24 & 3.14 & 3.45  \\ 
        
        SyncMVD~\cite{syncmvd}  & 26.96  & 5.37  & 23.08 & 4.01 & 4.12 & 3.71 \\ 
        \midrule
        Ours     & \textbf{20.02}  & \textbf{3.12}  & 23.25   & \textbf{4.13} & \textbf{4.51} & \textbf{4.21}  \\ 
        \bottomrule
       \end{tabular}
   }
\end{table}

\subsection{Text-Instructed 3D Texture Generation}
\label{text prompt}

\noindent
\textbf{Dataset Details.}
We filter out $104$k Objaverse~\cite{objaverse} samples with valid single texture maps and bake the texture with Xatlas~\cite{xatlas} wrapping, where $102$k samples are selected for training and the rest for validation and evaluation. 
The training data consists of rendered per-view RGB images $\mathbf{I}_{\text{N}}$ and their corresponding control proxy images $\mathbf{P}_{\text{N}}^l$.
The textual annotations are derived by instructing the CogVLM-2 model~\cite{cogvlm2} to describe category, texture, and appearance of the 3D objects, utilizing the multi-view images $\mathbf{I}_{\text{N}}$ as the model input.
Afterward, keywords are summarized using another LLM~\cite{mistral}.

\noindent \textbf{Evaluation Benchmarks.}
To comprehensively evaluate arbitrary types of meshes for the text-to-texture (T2T) generation, we build two evaluation benchmarks: 
\textbf{(1)} {the Objaverse T2T benchmark.}
To achieve a diverse Objaverse T2T benchmark, we integrate the Paint3D test set~\cite{paint3d}, containing 301 artist-crafted meshes from Objaverse, along with 3D scans and complex scenes, such as 3D character models, which leads to an extensive test set with a total of $1000$ models. 
\textbf{(2)} {the GSO T2T benchmark.}
The Google Scanned Objects (GSO) dataset provides a curated collection of 1032 3D scanned common household items, each captured at high resolution to capture intricate details. 
As 3D texture generation methods are primarily trained on 3D objects from Objaverse, the GSO dataset could be used to assess their generalizability.
Therefore, we establish the GSO T2T benchmark using the complete GSO dataset.

\noindent
\textbf{Evaluation metrics.} 
After generating 3D textures, $512$-resolution images of the mesh are rendered with the generated textures from 16 fixed viewpoints at the same elevation, \emph{i.e.}, $15$\textdegree~for fairness.
Then, we compare the rendered images with the true image distribution generated using the ground truth textures.
For a thorough evaluation, we use common generative metrics: Fréchet Inception Distance (FID)~\cite{fid}, Kernel Inception Distance (KID)~\cite{kid}, and CLIP score~\cite{clipscore} to assess image distribution, quality, and richness. 
The KID values are scaled by $10^3$ in all tables.

\noindent
\textbf{User Study.} 
To complement the quantitative results based on the generative metrics, we also conduct a user study to capture human preferences regarding the generated 3D textures.
$10$ participants are invited to evaluate the textured meshes in an interface that allows free navigation and observation of the 3D models. 
Each participant is asked to rate the following aspects on a scale of $1$ to $5$: overall quality, seam visibility, and overall consistency reflecting their preferences for each criterion.

\begin{figure}[t]
    \centering
    \includegraphics[width=\linewidth]{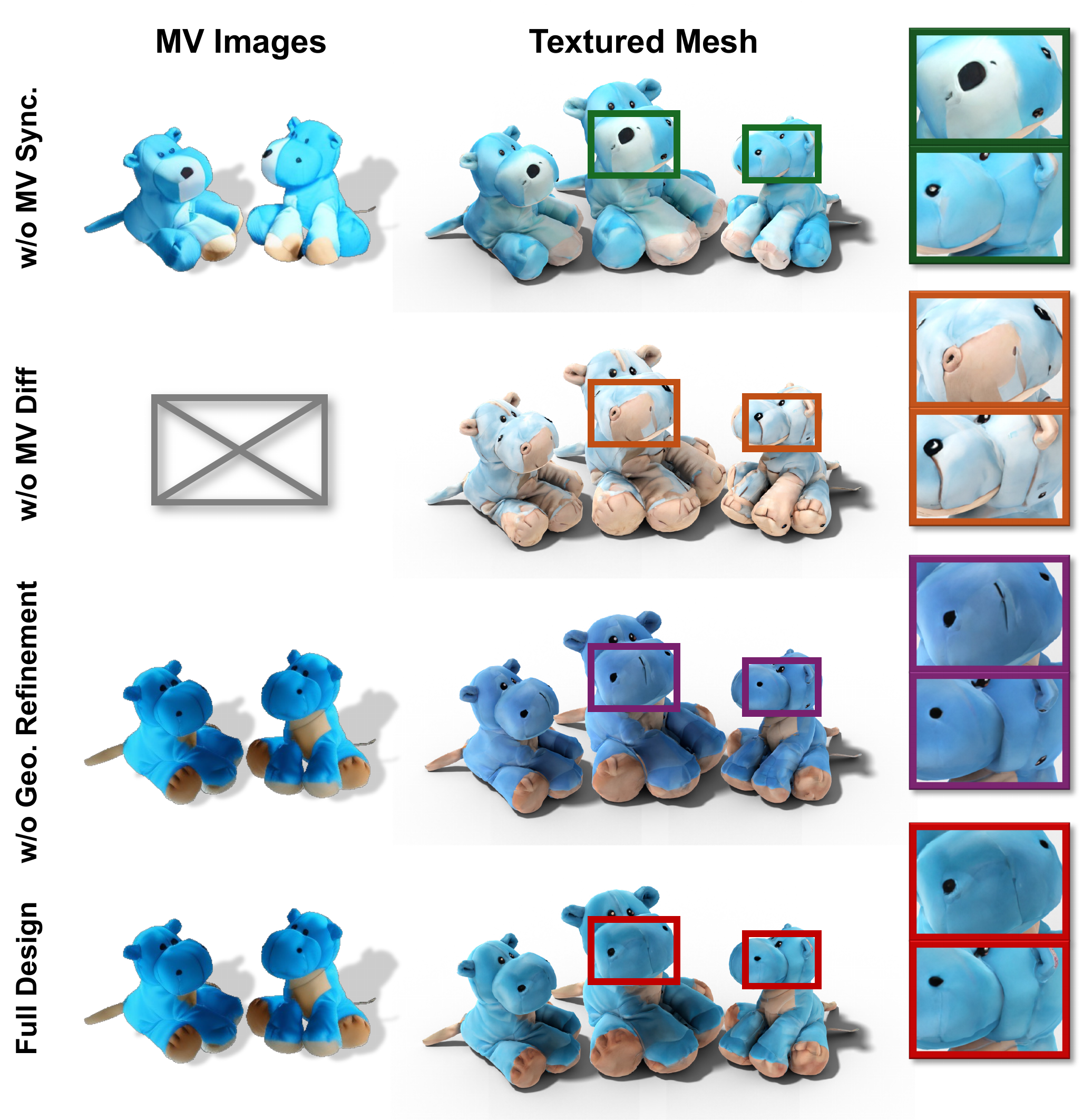}
    \vspace{-7mm}
    \caption{\textbf{Ablation Study on SMG Designs.} } 
    \label{fig:stage1_ablation}
\end{figure}

\begin{figure*}[t]
    \centering
    \vspace{-6mm}
    \includegraphics[width=0.95\linewidth]{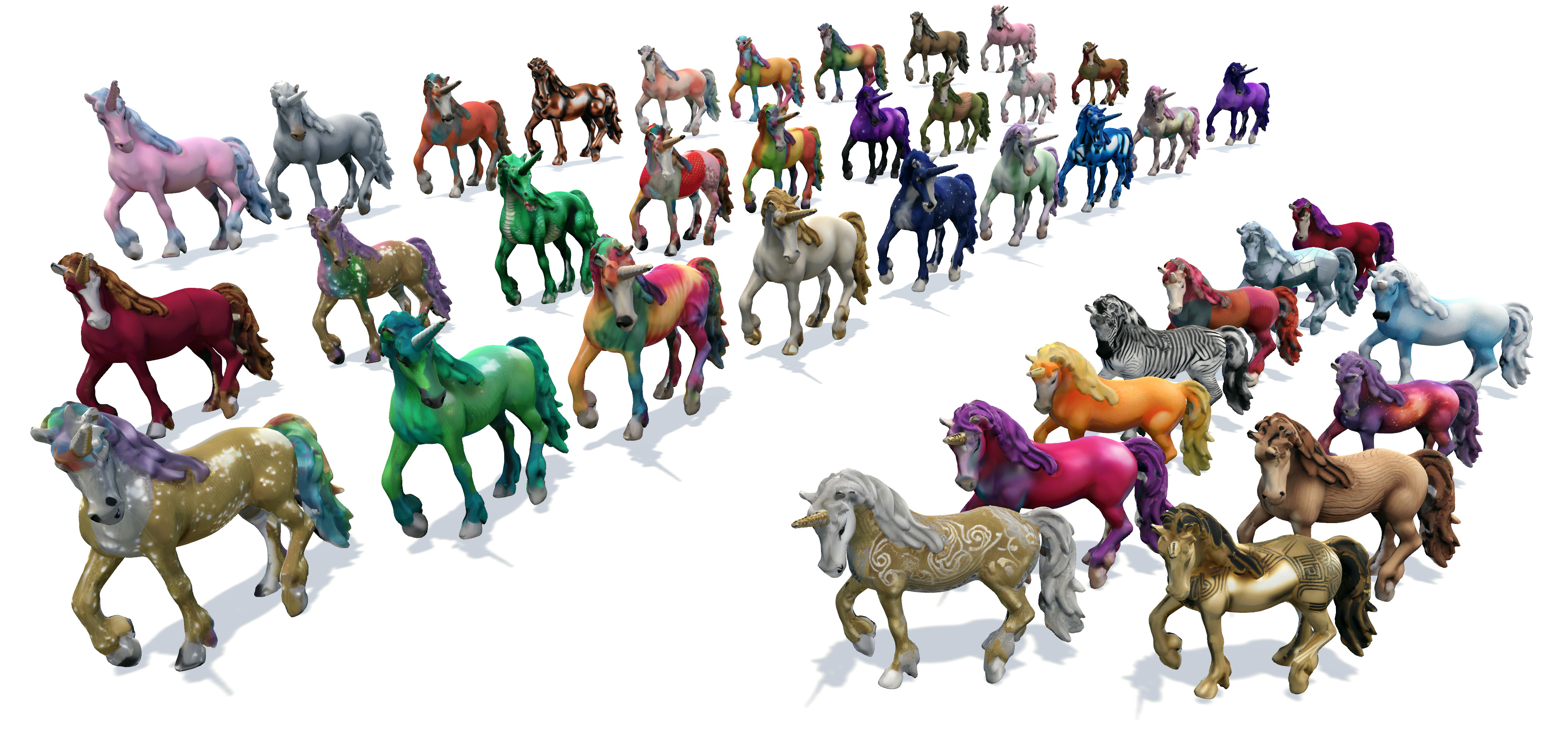}
    \vspace{-6mm}
    \caption{
    \textbf{Diverse Texturing on the Same Model.} 
    We use GPT4~\cite{gpt4} to generate 38 random texturing prompts without cherry-picking.}
    \label{fig:diversity}
\end{figure*}

\noindent
\textbf{Evaluation Results.}
We select all existing open-source SoTA methods on T2T for comparison, including TEXTure~\cite{Texture}, Paint3D~\cite{paint3d}, and SyncMVD~\cite{syncmvd}.
The quantitative results on the Objaverse T2T benchmark are reported in Tab.~\ref{tab:method_objaverse}. 
Our method, \OM{} achieves the best scores in terms of FID and KID, outperforming previous SoTA methods by over $4.3$ and $1.7$, respectively, while TEXTure offers the best CLIP score. 
It is worth noting that the TEXTure often encounters the Janus problem, potentially resulting in a high CLIP score.
According to the user study, \OM{} outperforms previous SoTA methods, achieving the highest ratings across all evaluated aspects, including overall quality, seam visibility, and consistency.

Since none of the T2T methods incorporate the GSO dataset in their training, the GSO T2T benchmark could be used to evaluate their generalizability.
The quantitative results on the GSO T2T benchmark are reported in Tab.~\ref{tab:method_gso}.  
Similar to the results observed in the Objaverse T2T benchmark, \OM{} achieves the best objective FID, KID performance, subjective user study scores, and second-best CLIP scores on the GSO T2T benchmark.
As Paint3D incorporates a crucial submodule that is trained on the Objaverse dataset, there is a noticeable decline in performance when evaluated on the GSO T2T benchmark, compared to its performance on the Objaverse T2T benchmark.
In contrast, \OM{} maintains its ability to produce high-quality 3D textures on the GSO benchmark, despite being trained on a dataset curated from Objaverse.
The qualitative comparisons could be visualized in Fig.~\ref{fig:qua}.

\begin{table}[]
    \setlength\tabcolsep{12pt}
    \caption{\textbf{Ablation Study on Objaverse~\cite{objaverse} Benchmark.} }
    \vspace{-3mm}
    \label{tab:ablation_main}
    \small
    \centering
    \begin{tabular}{l|ccc}
        \toprule
        \textbf{\fontsize{8}{7.5}\selectfont Method} & \textbf{\fontsize{8}{7.5}\selectfont FID$\downarrow$} & \textbf{\fontsize{8}{7.5}\selectfont KID$\downarrow$} & \textbf{\fontsize{8}{7.5}\selectfont CLIP$\uparrow$} \\ 
        \midrule
        \fontsize{7.5}{7.5}\selectfont w/o MV Sync.  & \fontsize{7.5}{7.5}\selectfont 21.42  & \fontsize{7.5}{7.5}\selectfont 3.72  & \fontsize{7.5}{7.5}\selectfont 19.90   \\ 
        \fontsize{7.5}{7.5}\selectfont w/o MV Diff  & \fontsize{7.5}{7.5}\selectfont 27.63  & \fontsize{7.5}{7.5}\selectfont 5.82  & \textbf{\fontsize{7.5}{7.5}\selectfont 20.44}    \\ 
        \fontsize{7.5}{7.5}\selectfont w/o Geo. Refinement  & \fontsize{7.5}{7.5}\selectfont 21.17  & \fontsize{7.5}{7.5}\selectfont 3.67  & \fontsize{7.5}{7.5}\selectfont 20.00    \\ 
        \hline
        \fontsize{7.5}{7.5}\selectfont w/o 3D Inpainting  & \fontsize{7.5}{7.5}\selectfont 20.91 & \fontsize{7.5}{7.5}\selectfont 3.56 & \fontsize{7.5}{7.5}\selectfont 19.87   \\ 
        \fontsize{7.5}{7.5}\selectfont w/o Seam Smoothing  & \textbf{\fontsize{7.5}{7.5}\selectfont 20.82} & \fontsize{7.5}{7.5}\selectfont 3.54 & {\fontsize{7.5}{7.5}\selectfont 19.92}  \\ 
        \midrule
        \fontsize{8}{7.5}\selectfont Full Design     & \fontsize{8}{7.5}\selectfont 20.89 & \textbf{\fontsize{8}{7.5}\selectfont 3.45} & \fontsize{8}{7.5}\selectfont 19.87  \\ 
        \bottomrule
    \end{tabular}
\end{table}

\subsection{Ablation}
\label{ablation}

\noindent
\textbf{SMG Designs.}
To validate the effectiveness of SMG module from Stage 1, we conduct a detailed ablation study over its three key designs, including the T2MV diffusion model, multi-view synchronization, and geometry-aware refinement.
Specifically, in \textbf{w/o MV Sync.} we omit the synchronization module and use $\mathbf{I}_{\mathrm{N}}^l$ rather than $\mathbf{I}^l_{\mathrm{N}_\text{sync}}$ for T2MV model. 
In \textbf{w/o MV Diff} we omit the whole T2MV modol $\mathcal{D}_\text{MV}$ and its output $\mathbf{I}^l_{\mathrm{N}_\text{sync}}$, and use the $\tau_g$ in the $\mathcal{D}_\text{I2I}$ to generate from scratch. 
In \textbf{w/o Geo. Refinement}, we drop the geometry refinement control model $\tau_g$ and only use the tiling model $\tau_t$. 
The quantitative results of the combined benchmarks are reported in Tab.~\ref{tab:ablation_main} which testify to the efficacy of the proposed designs. 

The qualitative results are illustrated in Fig.~\ref{fig:stage1_ablation}.
Without multi-view synchronization, inconsistent MV images result in poor initializations for subsequent refinement.
In the absence of MV images, the refinement network struggles to identify the orientation of the untextured mesh across different views, leading to the Janus problem.
Without geometry-guided refinement ($\tau_g$), the refinement merely adds details to coarse MV images and exacerbates initialization artifacts.
With the full design, \OM{} could produce multi-view consistent results, free of the Janus problem. 
Quantitative ablation results on the GSO benchmark are presented in Tab.\ref{tab:ablation_smg_gso}, with further analysis in Sec.~\ref{subsec: additional_results}.

\begin{figure}[]
    \centering
    \includegraphics[width=\linewidth]{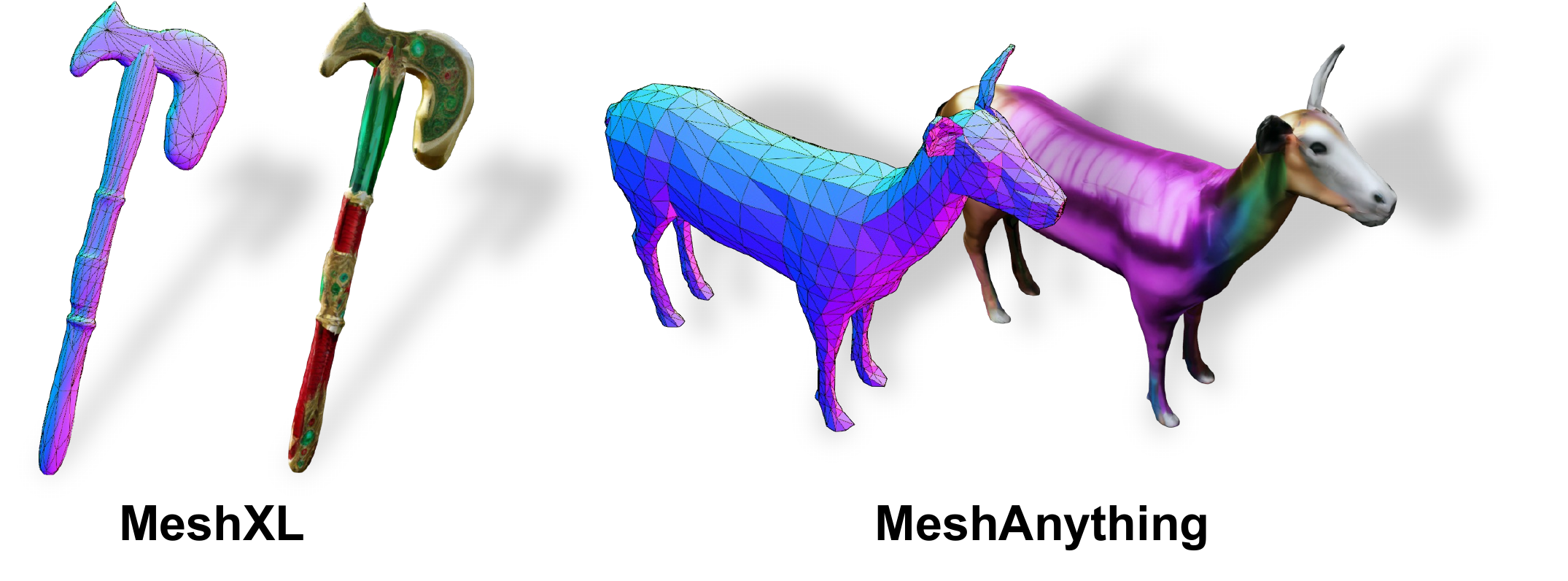}
    \vspace{-8mm}
    \caption{\textbf{Application of \OM.} Texturing for generated 3D assets from MeshXL~\cite{meshxl} and MeshAnything~\cite{meshanything,meshanythingv2}.}
    \label{fig:application}
\end{figure}

\noindent 
\textbf{3D Inpainting and Seam Smoothing.}
We also validate the effectiveness of 3D inpainting (from Stage 2) and seam smoothing (from Stage 3), and the quantitative results are reported in Tab.~\ref{tab:ablation_main}. 
Their quantitative results are very close, with the FID showing slight variations within $\pm0.1$ and the KID consistently remaining below $0.2$.
Although these refinement operations have minimal impact on metrics, they could effectively eliminate texture artifacts.

\subsection{Application}

\OM{} could generate faithful 3D textures based on text instructions, hence supporting various related applications, such as: 
\textbf{1) Generating Diverse 3D Texture}: 
Given a certain 3D mesh model, \OM{} could generate 3D textures with large variations conditioned on different text prompts.
As shown in Fig.~\ref{fig:diversity}, \OM{} generates $38$ different 3D textures for the single unicorn model.
\textbf{2) Texturing AI-Generated 3D Meshes}: 
\OM{} generates 3D textures independently of the UV unwrapping quality, enabling it to produce high-quality textures even when the AI-generated 3D meshes contain minor artifacts. 
We demonstrate the textures generated by \OM{} on 3D meshes created by MeshXL~\cite{meshxl} and MeshAnything~\cite{meshanything,meshanythingv2} in Fig.~\ref{fig:application}.
\section{Conclusion}

In this paper, we introduce \textsf{\textcolor{citypink}{MVP}\textcolor{cityblue}{aint}}, a comprehensive framework for generating 3D textures from text, consisting of three key stages: synchronized multi-view generation, 3D space texture inpainting, and UV refinement.
Utilizing synchronized multi-view diffusion, \OM{} initializes 3D textures based on generated multi-view images, ensuring high cross-view consistency.
Subsequently, areas not covered by these multi-view images are textured through inpainting in 3D space. 
Finally, a refinement module enhances and upscales the 3D mesh in the UV space, producing high-quality UV textures at a 2K resolution.
Extensive experiments demonstrate that \OM{} consistently produces high-quality 3D textures, outperforming existing SoTA texturing methods.

{
    \small
    \balance
    \bibliographystyle{ieeenat_fullname}
    \bibliography{cvpr}
}


\clearpage
\appendix

\noindent
\textbf{\LARGE Appendix}
\vspace{5ex}

\setcounter{equation}{0}
\setcounter{figure}{0}
\setcounter{table}{0}
\setcounter{section}{0}
\makeatletter
\renewcommand{\theequation}{S\arabic{equation}}
\renewcommand{\thefigure}{S\arabic{figure}}
\renewcommand{\thetable}{S\arabic{table}}

\section{Detailed Discussion on Related Works}\label{sec:detailed_related}
\subsection{Baseline Methods}
In this section, we review the most representative baseline that we compared during the evaluation and discuss its strengths and weaknesses. Other great close-source methods, such as Meta 3D TextureGen~\cite{meta3dtexgen}, are removed from the comparison scoop. We omit the SOTA method Text2Tex~\cite{text2tex} in this report due to the extensive evaluation in previous literature~\cite{paint3d,syncmvd}.

\noindent \textbf{TEXTure. } TEXTure~\cite{Texture} presents a method for generating 3D textures from textual descriptions using a pretrained depth-to-image diffusion model. TEXTure employs an iterative approach to ensure consistent texturing from multiple viewpoints by dividing rendered images into ``keep'', ``refine'', and ``generate'' regions. It supports texture transfer and editing through both text prompts and user input. However, the method can produce global inconsistencies when handling complex geometries or viewpoints that do not fully capture the model, which the authors identify as areas for future improvement.
 
\noindent \textbf{Paint3D. } Paint3D~\cite{paint3d} is a two-stage generative method. In the first stage, it utilizes ControlNet to generate textures from individual viewpoints. In the second stage, it innovatively proposes direct texture generation in UV space. Paint3D~\cite{paint3d} employs a UV position map as the control signal to train a ControlNet~\cite{controlnet}, leveraging the generative capabilities of existing diffusion models to produce corresponding textures. However, due to significant differences between texture map samples and image samples, the diffusion network inherently lacks robust texture generation capabilities. Moreover, during the control generation process using the UV position map, the diffusion model tends to assign similar colors to atlas textures that are closer in 2D mapping space, rather than to the 3D space represented by the control channels. As a result, Paint3D~\cite{paint3d} performs poorly on most automatically unwrapped complex texture maps, leading to numerous instances of misaligned textures.

\noindent \textbf{SyncMVD. }
SyncMVD \cite{syncmvd} proposes a zero-shot texture generation method. It addresses the limitations of asynchronous diffusion that plague traditional project-and-inpaint techniques. While previous approaches generate textures from individual views without adequate synchronization, leading to inconsistencies, 
SyncMVD synchronizes the diffusion processes for multi-view generation.
This innovative method facilitates early consensus in texture generation by sharing denoised content across overlapping views during each denoising step. 
As a result, SyncMVD~\cite{syncmvd} achieves consistent textures that exhibit remarkable details and coherence across various perspectives. 
We highlight that SyncMVD does not utilize a multi-view diffusion model for generating multi-view images.
Instead, SyncMVD retains the iterative camera pose and subject framework generated by single-view diffusion, which often leads to multi-faces problems due to the lack of constraints between multiple views. 

\subsection{Border Related Methods}
In this section, we discuss the other related works that share similar designs that are out of our evaluation's scope.

\noindent \textbf{Meta 3D TextureGen. }
Meta 3D TextureGen \citep{meta3dtexgen} proposes a two-stage texture generation method. In the first stage, normal and position controls are used to generate aesthetically pleasing multi-view images, which are then weighted and projected based on the mesh surface normals and view directions to obtain an initial UV texture map. In the second stage, the process continues in the UV space, where normal and position controls are again employed for inpainting. Finally, texture enhancement is applied to perform super-resolution and enrich texture details. Through this innovative approach, Meta 3D TextureGen can generate high-definition textures that are rich in detail and aesthetically pleasing with Emu~\cite{emu} base model. However, due to the relatively independent multi-view generation in the first stage, there may be a lack of consistency between the textures across different views and may be potentially vulnerable to the Janus problem. Additionally, the limited number of viewpoints in the first stage can lead to significant unobserved areas for objects with complex occlusions. This poses challenges for the subsequent texture inpainting and enhancement processes, which are performed entirely in UV space. Since the code for Meta 3D TextureGen has not been open-sourced, we do not include it in our experimental comparisons.

\noindent \textbf{Unique3D.}
Unique3D~\cite{unique3d} is an innovative image-to-3D framework capable of efficiently generating high-quality 3D meshes from single-view images, demonstrating both high fidelity and robust generalization capabilities. It integrates multi-view diffusion models, multi-scale upsampling strategies, and the proposed ISOMER algorithm, enabling the generation of detail-rich textured meshes in 30 seconds. Similar to our method, Unique3D uses a coarse-to-fine manner to generate multi-view high-resolution images. Differently, Unique3D targets simultaneous geometry and texture generation, and 3D consistency is encouraged but not strictly aligned. For example, it upscales $256\times256$ images with a multi-view ControlNet~\cite{controlnet} and further boosts the resolution to $2$k with a view-separate upscaling model without any synchronization.

\noindent \textbf{CLAY.}
CLAY~\cite{clay} is a large-scale 3D generation model that can produce high-quality 3D geometry and materials from text or image inputs. It employs a multi-resolution VAE and a minimal latent diffusion transformer, supporting various control modalities such as multi-view images and voxels to facilitate precise 3D asset creation. The texturing module of CLAY~\cite{clay} is also built upon MVDream~\cite{mvdream}, they modify it by adding additional channels and modalities to support physical-based rendering~(PBR), ControlNet~\cite{controlnet} to achieve view control, and uses LoRA~\cite{lora}-based fine-tuning. CLAY~\cite{clay} generates the 4 orthogonal views of images, and similar to Paint3D~\cite{paint3d} inpaint and upscale the PBR images in UV space. We argued inpainting directly on UV space is vulnerable to complicated UV unwrapping, especially since the coarse UV is generated from a low coverage of 4 views at a resolution of $256\times256$ with great loss in detail.
Note that CLAY focuses on PBR generation, and it has the image-prompt ability, whereas we more focus on text-to-texture generation. As the implementation is not publicly available, we exclude CLAY in our evaluation.

\section{Implementation Details}
\label{sec:implementation_details}

\subsection{Control Strength of I2I Finer Painting} 
\begin{figure*}
    \centering
    \includegraphics[width=0.94\linewidth]{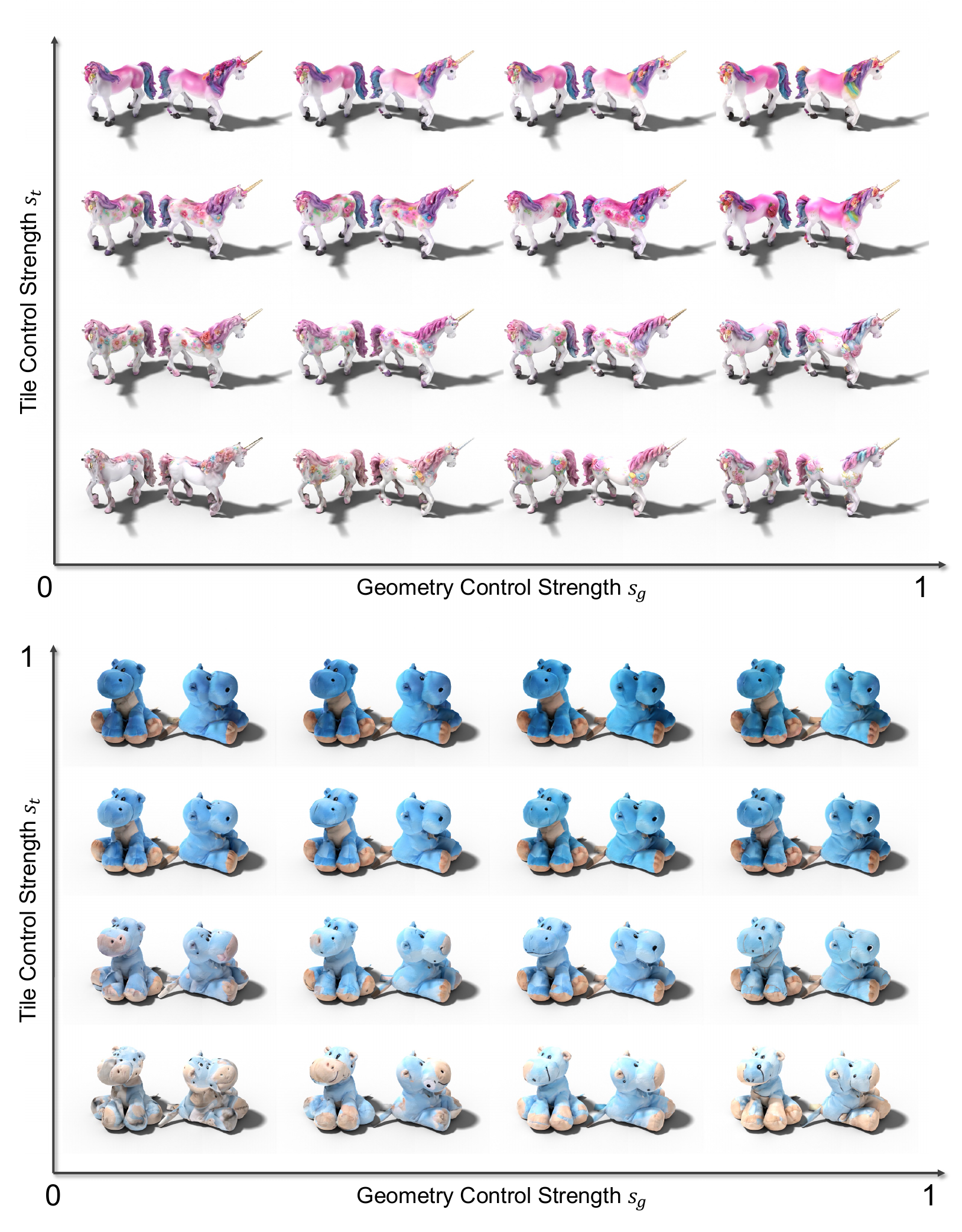}
    \vspace{-4.5mm}
    \caption{\textbf{High Flexibility in Proposed SMG Design.} With two refinement modules $\tau_g, \tau_t$ and their corresponding control strength $s_g, s_t$, \OM~can provide user versatile choices of texture generation. With a larger strength of $s_t$, the generated results will have more alignment with coarse MV images. With a larger strength of $s_g$, the generated results will have more creativity (see the `Unicorn' case) while with a higher risk of Janus problem (see the `Blue Hippo' case).}
    \label{fig:i2i}
\end{figure*}

In the Synchronized Multi-view Generation (SMG) process, we employed a refinement model composed of two control modules. The first module, denoted as $\tau_t$, is designed to provide low-resolution multi-view (MV) images that are free of Janus artifacts and maintain multi-view consistency for the refinement stage. The second module, $\tau_g$, offers geometric guidance, ensuring that the multi-view refinement process is fully aligned with the underlying mesh structure. These two components are controlled by the parameters $s_t$ and $s_g$, which allow for flexible user-defined configurations. If the user is satisfied with the generated MV images, a relatively high $s_t$ can be set to maintain consistency, while $s_g$ can be reduced to minimize divergence, as demonstrated in the case of the `Blue Hippo' in the Fig.~\ref{fig:i2i}. Conversely, if the user is dissatisfied with the MV images or desires more creative diversity, a lower $s_t$ and higher $s_g$ can be used, as shown in the `Unicorn' case. It is important to note that without any $s_t$, our refinement model may face Janus issues similar to those observed in SyncMVD~\cite{syncmvd} and TEXTure~\cite{Texture} illustrated by the `Blue Hippo' case in Fig.~\ref{fig:i2i}.

\subsection{Spatial-aware 3D Inpainting} \label{subsec: S3I}
In Sec.~\ref{Stage Two}, we briefly described the execution flow of the algorithm. Here, we provide a more detailed explanation of the algorithm. \textbf{Spatial-aware 3D Inpainting (S3I)} is designed to inpaint unobserved regions after multi-view projection. We sample the mesh into a dense point cloud, enabling us to apply coloring at the point cloud level to capture structural information. S3I is a learning-free method that propagates color from observed regions to unobserved regions.

Since the propagation is based on the k-nearest neighbors (KNN) algorithm, there can be errors in propagation across planes, particularly in seam areas. To address this, we calculate the normal vector for each point, incorporating it into the weight calculation to prevent color diffusion across non-coplanar surfaces. The pseudocode for the detailed algorithm execution is as follows: 

\begin{algorithm}[t]
\caption{Spatial-aware 3D Inpainting}
\KwIn{
\textit{colored\_points}: A set of points with color information\\
\textit{color\_mask}: boolean array where \textit{true} indicates points with valid color\\
\textit{n}: number of nearest neighbors for KNN search
}
\KwOut{\textit{updated\_colored\_points}: A set of points with updated color information}

\SetKwFunction{FMain}{update\_colored\_points}
\SetKwFunction{FKNN}{knn\_color\_completion}
\SetKwProg{Fn}{Function}{:}{}

\Fn{\FMain{\textit{colored\_points}, \textit{colored\_mask}}}{
    
    \textit{points} $\gets$ \textit{colored\_points.points}\\
    \textit{colors} $\gets$ \textit{colored\_points}[\textit{color\_mask}].\textit{colors}\\
    \textit{normals} $\gets$ calculate surface normals of \textit{points}\\
    
    \textit{unknown\_points} $\gets$ \textit{points}[\textit{color\_mask}]\\
    \textit{unknown\_normals} $\gets$ \textit{normals}[\textit{color\_mask}]\\
    
    \textit{tree} $\gets$ KDTree(\textit{points}, \textit{n})\\
    \textit{distances, indices} $\gets$ \textit{tree}(unknown\_points)\\

    \textit{neighbors\_normals} $\gets$ Index\_Select(normals, indices)\\
    \textit{cos} $\gets$ Cosine\_Similarity(\textit{unknown\_normals}, \textit{neighbors\_normals})\\
    
    \textit{distance\_score} $\gets$ Normalize(1 / \textit{distances}) \\
    \textit{weight} $\gets$ \textit{cos} * \textit{distance\_score} \\

    \textit{coloring round} $\gets$ 0\\
    \While{\textit{stage} == \textit{"uncolored"} or \textit{coloring\_round} $>$ 0}{
        \For{\textit{point} in \textit{unknown\_points}}{
            \textit{neighbors} $\gets$ KNN(\textit{points}, \textit{n})\\
            \textit{new\_color} $\gets$ Weighted-Average(\textit{weight}, \textit{neighbors)}\\
            \textit{colored\_points}.\textit{assign\_color}(\textit{point}, \textit{new\_color})\\
        }

        Calculate total number of colored points\\
        
        \If{coloring progress}{
            Increment coloring round\\
        }
        \Else{
            Decrease coloring round or exit loop if no further progress\\
        }
    }

    \Return \textit{colored\_points}\\
}

\end{algorithm}

\subsection{Spatial-Aware Seam-Smoothing Algorithm.} \label{subsec: SSA}
In Sec.~\ref{Stage Three}, we introduce the implementation concept of the \textbf{Spatial-Aware Seam-Smoothing} algorithm. It is used to correct color discontinuities at seams after super-resolution in the UV space. Similar to S3I, it employs k-nearest neighbor (KNN) search to perform weighted color averaging. The detailed pseudocode for the algorithm is as follows.

\begin{algorithm}[t]
\caption{KNN Seam Smoothing Algorithm}
\KwIn{
    \textit{colored\_points}: A set of points with color information, where each row is $[xyzrgb]$ \\
    \textit{seam\_mask}: boolean array where true indicates seam points \\
    \textit{n}: number of nearest neighbors for KNN search
}
\KwOut{
    \textit{new\_color}: Smoothed color for seam points
}

\SetKwFunction{FMain}{knn\_seam\_smooth}
\SetKwProg{Fn}{Function}{:}{}
\Fn{\FMain{colored\_points, seam\_mask, n\_neighbors}}{
    \textit{non\_seam\_points} $\gets$ \textit{colored\_points}[$\neg$seam\_mask]
    \textit{normals} $\gets$ calculate surface normals of \textit{colored\_points} \\
    \textit{seam\_normals} $\gets$ \textit{normals}[seam\_mask] \\
    \textit{non\_seam\_normals} $\gets$ \textit{normals}[$\neg$seam\_mask] \\
    \textit{colors} $\gets$ \textit{non\_seam\_points.colors} \\
    \textit{tree} $\gets$ KDTree(\textit{non\_seam\_points}, \textit{n})\\
    \textit{distances, indices} $\gets$ \textit{tree}(\textit{seam\_points})\\
    \textit{seam\_neighbors\_normals} $\gets$ Index\_Select(\textit{non\_seam\_normals}, \textit{indices}) \\
    \textit{cos} $\gets$ Cosine\_Similarity(\textit{seam\_normals}, \textit{seam\_neighbors\_normals}) \\
    \textit{distance\_score} $\gets$ Normalize(1 / \textit{distances}) \\
    \textit{weight} $\gets$ \textit{cos} * \textit{distance\_score} \\
    
    \For{\textit{point} in \textit{seam\_points}}{
            \textit{neighbors} $\gets$ KNN(\textit{colored\_points}, \textit{n})\\
            \textit{new\_color} $\gets$ Weighted-Average(\textit{weight}, \textit{neighbors.colors)}\\
            \textit{colored\_points}.\textit{assign\_color}(\textit{point}, \textit{new\_color})\\
        }
    \Return \textit{colored\_points}.\textit{colors}
}

\end{algorithm}

\subsection{Unprojection Reduction Algorithm} \label{subsec: URA}
\begin{figure}
    \centering
    \includegraphics[width=0.9\linewidth]{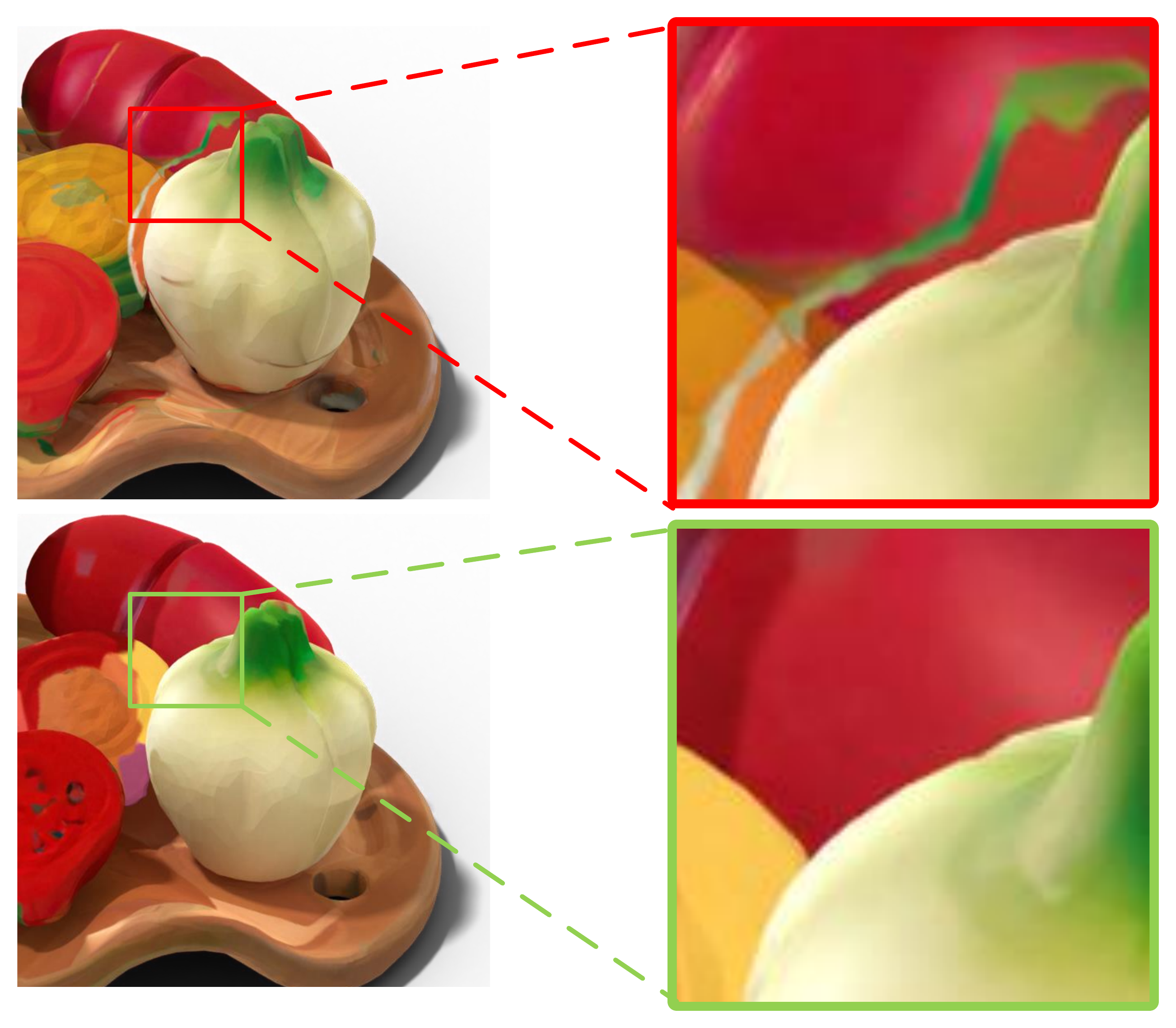}
    \caption{\textbf{Illustration of Projection Error}.
    When the Unprojection Reduction Algorithm is not used, occlusions may cause the color of the occluding object to be projected onto the occluded areas, resulting in artifacts. However, by applying the Unprojection Reduction Algorithm, this issue can be effectively resolved by preventing incorrect color projections onto occluded regions.}
    \label{fig:projection}
\end{figure}

After generating consistent multi-view images, we apply weighted projection to obtain the texture UV map. However, due to occlusion in 3D objects, this can lead to projection errors and artifacts in the UV map, as shown in Figure~\ref{fig:projection}. To address this, we determine occlusion relationships and reduced the affected projection areas. The specific algorithm workflow is as follows: we first extract the 3D coordinates for each valid pixel in the UV map, generating a 3D point cloud. Then, using the occlusion detection algorithm proposed in \citep{projection_issue}, we mark occluded points for each view. Finally, these occlusion marks are mapped back onto the UV map, and regions marked as "occluded" are excluded from projection.

\subsection{Discussion on UV Space Tiling}\label{subsec:UV_tiling}
Depending on the specific generation requirements, our UV upscale model $\mathcal{D}_\text{UP}$ can be replaced by a refinement network $\mathcal{D}_\text{TILE}$. Similar to the tiling network in the second stages in Paint3D~\cite{paint3d} and Meta 3D TextureGen~\cite{meta3dtexgen}, $\mathcal{D}_\text{TILE}$ is a diffusion model controlled by a tiling control module $\tau_\text{tile}$ and a position map control module $\tau_\text{pos}$. The position map is a UV map $\mathbf{T}_\text{pose}$ where the 3D position of the corresponding mesh surface replaces the channel values. Formally, this optional module can be written as 

\begin{equation}
    \mathbf{T}_\text{TILE} = \mathcal{D}_\text{TILE}(z^\text{UV}, \mathbf{T}_\text{c}, \mathbf{T}_\text{pose}; \tau_\text{tile}, \tau_\text{pos}).
\end{equation}

\begin{figure}[b!]
    \centering
    \includegraphics[width=\linewidth]{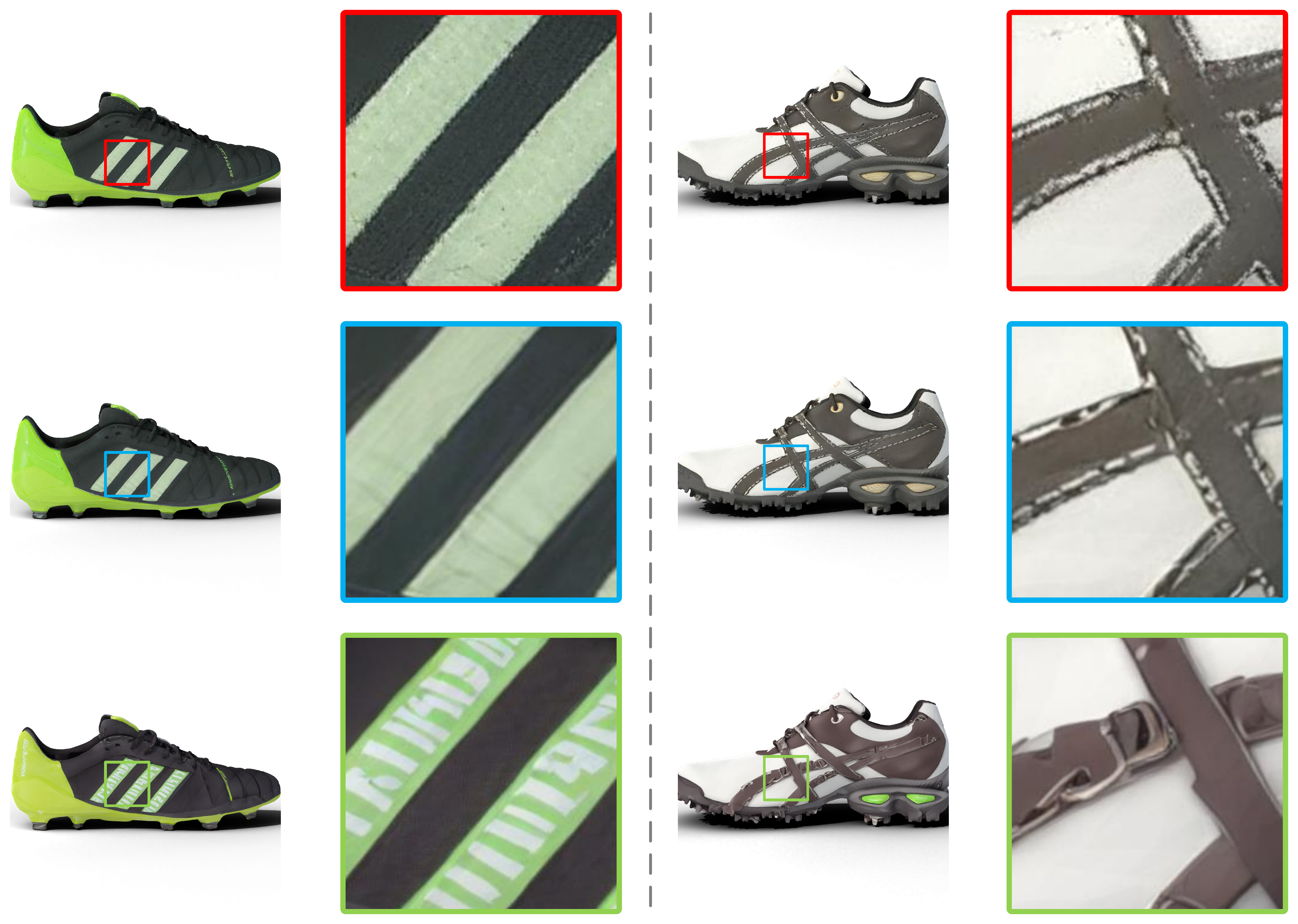}
    \caption{\textbf{Discussion on UV upscale or tiling.} Given groundtruth low-resolution UV map (first row), UV upscaling $\mathcal{D}_\text{UP}$ generates sharp and clean textures (second row), UV tiling $\mathcal{D}_\text{TILE}$ will add intricate but acceptable details (the left case in the third row) or wrong details (the right case in the third row).}
    \label{fig:discussion_uv_tiling}
\end{figure}

Different from previous literature~\cite{paint3d,meta3dtexgen}, we find the UV tiling process is very vulnerable to continuity or the UV wrapping, leading to obvious seams on 3D mesh when UV atlas is packed randomly like Xatlas~\cite{xatlas}. Thus, we only treat this module as an opt for $\mathcal{D}_\text{UP}$ if users want to pursue extreme details. We give an example of a comparison of UV upscaling and tiling in Fig.~\ref{fig:discussion_uv_tiling}, where tiling can increase the upper bound of generation quality, while still suffering from stability. Thus, we use $\mathcal{D}_\text{UP}$ as our default super-resolution model, and it is very important to point out that all the results in this work are generated by $\mathcal{D}_\text{UP}$.

\begin{figure*}[ht]
    \centering
    \includegraphics[width=0.95\linewidth]{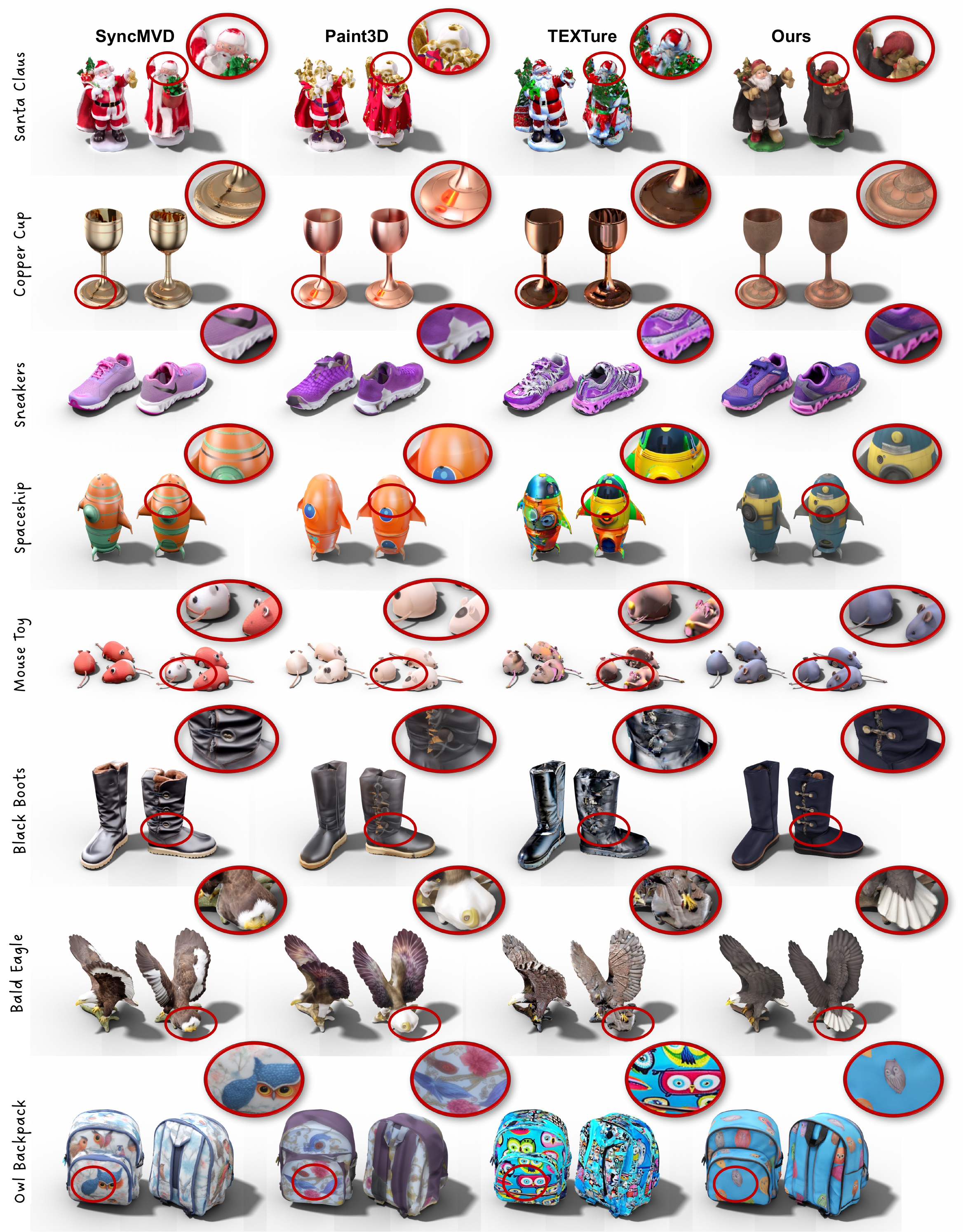}
    \caption{\textbf{Additional results with border categories.} The 3D models are from GSO~\cite{gso} and Objaverse~\cite{objaverse}, and the text prompt is abbreviated.} 
    \label{fig:qualitative-compair}
\end{figure*}

\section{Detailed Evaluations}
\label{sec:detailed_evaluations}

\subsection{Detailed Evaluation Settings}\label{sec:detailed-exp-setting}

\noindent
\textbf{Evaluation Elevation Selection.}
During the comparative experiments, we observed that each method predefined its optimal elevation angle, complicating the selection of a unified rendering perspective for comparison. Our approach involved setting the elevation angle for Paint3D and TEXTure to $30$\textdegree, while SyncMVD utilized a rendering perspective comprising eight evenly distributed views at $0$\textdegree and two views at $\pm60$\textdegree, which we retained. Our proposed method also employed four views at $30$\textdegree and four views at $-30$\textdegree. Consequently, we selected $15$\textdegree~as the testing elevation and densely rendered 16 perspectives to ensure the validity of the evaluation. We also conduct additional ablation study on such setting design in Sec.~\ref{sec:ablation_view}.

\subsection{Quantitative and Qualitative Results}\label{subsec: additional_results}

\vspace{2mm}
\noindent \textbf{Additional Qualitative Results of T2T Evaluation.}
We provide additional qualitative results to showcase MVPaint's outcomes, including extra comparisons with baseline methods on border categories in Fig.~\ref{fig:qualitative-compair}). From the qualitative results, we can conclude that SyncMVD~\cite{syncmvd} can generate good results on general lifeless objects like shoes, bags, etc. While it has severe Janus problems on generating objects with heads, see the Santa, mouse, and eagle cases. Paint3D~\cite{paint3d}'s performance is highly related to UV wrapping or texture complexity. When given easy prompts are given like `copper cup' or `black boots', Paint3D~\cite{paint3d} can produce decent results when texture complexity is high like `Santa' or `eagle' it generates results with artifacts and seams. TEXTure~\cite{Texture} generally generates texture with high image saturation with large-scale artifacts and severe Janus problems, which accords with the user study in Tab.~\ref{tab:method_objaverse} and Tab.\ref{tab:method_gso} where TEXTure~\cite{Texture} has the least user appealing scores, even though it has good subject evaluation results.

\vspace{2mm}
\noindent\textbf{Quantitative Results of Ablation on SMG Designs.}
We also report the quantitative results of the ablation study on SMG designs on GSO benchmark in Tab.~\ref{tab:ablation_main}. Different from the results in in-domain analysis in Tab.~\ref{tab:ablation_main} where `\textbf{w/o MV Diff}' influence the most to the quantitative results, in cross-domain benchmark which mostly consists of scan objects, the influence of each design is much more even. What accords with the Objaverse~\cite{objaverse} benchmark results is that the full design achieves the best objective metrics and subjective metrics except for CLIP scores.

\begin{table}[ht]
   \caption{\textbf{Quantitative Results on SMG Designs on GSO Benchmark.}}
   \vspace{-3mm}
   \label{tab:ablation_smg_gso}
   \small
   \centering
   \resizebox{0.48\textwidth}{!}{
       \begin{tabular}{l|cccccc}
       \toprule
       \multirow{2}{*}{\textbf{Method}} & \multirow{2}{*}{\textbf{FID$\downarrow$}} & \multirow{2}{*}{\textbf{KID$\downarrow$}} & \multirow{2}{*}{\textbf{CLIP$\uparrow$}} & \multicolumn{3}{c}{\textbf{User Study}} \\
        & & & & \textbf{Overall $\uparrow$} & \textbf{Seamless$\uparrow$} & \textbf{Consistency$\uparrow$} \\ 
       \midrule
        w/o MV Sync.  & 25.72  & 5.31  & \textbf{22.49}  & 3.98 & 4.19 & 4.04   \\ 
        w/o MV Paint  & 25.23  & 5.17  & 23.36  & 4.04 & 3.99 & 3.84   \\ 
        w/o Geo. Refinement  & 25.56  & 5.27  & \textbf{22.49}  & 3.85 & 3.87 & 4.14   \\ 
        \midrule
        Full Design     & \textbf{20.02}  & \textbf{3.12}  & 23.25  & \textbf{4.13} & \textbf{4.51} & \textbf{4.21}  \\ 
        \bottomrule
       \end{tabular}
   }
\end{table}

\section{Additional Experiments}
\label{sec:additional_exp}

\subsection{Ablation on View Selection}\label{sec:ablation_view}
We conducted ablation experiments on the choice of the number of viewpoints based on the main paper's ablation study. Specifically, we tested several configurations: \(N=4\) with elevations $\phi=0$\textdegree; \(N=8\) with elevation $\phi=0$\textdegree; \(N=16\) with elevation at $\phi=0$\textdegree; \(N=8\) with elevations interleaved $\phi=\pm30$\textdegree; and \(N=16\) with elevations interleaved between $\phi=\pm30$\textdegree. All the view azimuths $\theta$ are uniformly distributed in all experiments. Our experiments were conducted on the Objaverse~\cite{objaverse} dataset, we use the testing elevation $\phi=15$\textdegree~and evaluation metrics. We organize quantitative results and report in Tab.~\ref{tab:ablation_view}. From the results, we observe that interleaved elevations at $\phi=\pm30$\textdegree~ yield better viewpoint coverage, achieving improved metrics at a novel elevation of $15$\textdegree. Additionally, we found that increasing \(N\) from 4 to 8 results in significant metric improvements, while further increasing to 16 leads to a decline in metrics. This is attributed to the high number of independently generated viewpoints during the refinement phase, which causes excessive overlap and leads to over-smoothed or blurry textures, resulting in lower performance metrics.
\begin{table}[h!]
   \caption{\textbf{Quantitative Results on View Number on Objaverse~\cite{objaverse} Benchmark.} }
   \vspace{-1mm}
   \label{tab:ablation_view}
   \small
   \centering
   \resizebox{0.4\textwidth}{!}{
       \begin{tabular}{l|ccc}
       \toprule
       \textbf{View Setting} & \textbf{FID$\downarrow$} & \textbf{KID$\downarrow$} & \textbf{CLIP$\uparrow$} \\ 
       \midrule
        $N=4, \phi=0$\textdegree & 35.48 & 11.24 & 23.21 \\
        $N=8, \phi=0$\textdegree & 23.45 & 4.12  & 20.48 \\
        $N=16, \phi=0$\textdegree & 25.71 & 4.51  & 21.45 \\
        \midrule
        $N=8, \phi=\pm30$\textdegree & \textbf{20.89} & \textbf{3.45}  & \textbf{19.87} \\
        $N=16, \phi=\pm30$\textdegree & 21.58 & 3.87  & 20.21 \\
        \bottomrule
       \end{tabular}
   }
\end{table}

\subsection{Runtime Evaluation}
We conducted runtime tests to comprehensively evaluate the proposed method's performance. Initially, we performed a detailed analysis of the average time required for each module, as illustrated in Tab.~\ref{tab:runtime_ours}. The total runtime of our pipeline was measured at $97.79$ seconds, with the majority of the computational load concentrated in the multi-view refinement step, which accounted for nearly $80\%$ of the overall runtime with SDXL~\cite{sdxl} model. 

\begin{table}[h!]
   \caption{
   \textbf{Runtime Evaluation of \OM~pipeline.}
   } 
   \vspace{-3mm}
   \label{tab:runtime_ours}
   \small
   \centering
   \resizebox{0.48\textwidth}{!}{
       \begin{tabular}{l|ccccc|c}
       \toprule
        \multirow{2}{*}{\textbf{Stage}} & \multicolumn{2}{c}{\textbf{I. SMG} } & \multirow{2}{*}{\textbf{II. S3I}} & \multicolumn{2}{c|}{\textbf{III. UVR}} & \multirow{2}{*}{\textbf{Overall}} \\
        & T2MV & I2I & & SR & Fix Seams\\
        \midrule
        Time (s) & 10.51 & 78.04 & 0.82 & 8.30 & 0.12 & 97.79 \\
        \bottomrule
       \end{tabular}
   }

\end{table}

A runtime comparison of our method against several SOTA approaches is also conducted. Experiments were executed on a uniform hardware environment utilizing an single H800 GPU, ensuring consistency in testing conditions. Notably, the model loading time and the rendering time for intermediate results or video outputs were excluded from the reported statistics. As illustrated in the following Tab.~\ref{tab:runtime_compare}, despite employing a greater number of steps and higher multi-view resolutions, our runtime remains on par with SOTA methods.

\begin{table}[h!]
   \caption{
   \textbf{Runtime Comparison between the \OM~and SOTA methods.}
   } 
   \vspace{-3mm}
   \label{tab:runtime_compare}
   \small
   \centering
   \resizebox{0.48\textwidth}{!}{
       \begin{tabular}{l|ccc|c}
       \toprule
        Methods & TEXTure~\cite{Texture} & Paint3D~\cite{paint3d} & SyncMVD~\cite{syncmvd} & \OM~ \\
        \midrule
        Time (s) $\downarrow$ & 142.52 & 104.36 & \textbf{87.32} & 97.79 \\
        \bottomrule
       \end{tabular}
   }

\end{table}

\section{Limitations}
\label{sec:limitation}
MVPaint achieves strong spatial consistency and high-resolution textures, but it also has certain limitations, primarily in three areas: the aesthetic problem and the lack of image prompt ability.

\vspace{2mm}
\noindent
\textbf{Aesthetic Problem.}
Compared to Meta 3D TextureGen~\cite{meta3dtexgen}, our model exhibits certain deficiencies in the aesthetic quality of texture generation. This can be attributed partly to Meta 3D TextureGen's~\cite{meta3dtexgen} use of an aesthetically optimized diffusion model, Emu~\cite{emu}, which generates each viewpoint independently. While this approach ensures aesthetic appeal, it may compromise the consistency of the textures. In contrast, our chosen T2MV model, MVDream~\cite{mvdream}, is based on SD2~\cite{sd2}, which lacks diversity and aesthetic quality in its texture outputs. To address this limitation, we propose the I2I refinement method, which employs a higher-resolution and more aesthetically pleasing model, SDXL~\cite{sdxl}, to supplement details or re-render the multi-view images. This significantly enhances the aesthetic quality. However, increasing the re-rendering intensity raises the probability of encountering the Janus Problem, thereby limiting our ability to generate highly aesthetically pleasing textures. We believe that the emergence of superior T2MV models in the future will help mitigate the aesthetic shortcomings of our current prototype.

\vspace{2mm}
\noindent
\textbf{Lack of Image Prompt Capability.}
Similar to Meta 3D TextureGen~\cite{meta3dtexgen} and SyncMVD~\cite{syncmvd}, our MVPaint focuses on the generation of textures from text, without emphasizing the optimization of input diversity; consequently, we do not address image prompts in this paper. In contrast, some single-view methods, such as TEXTure~\cite{Texture} and Paint3D~\cite{paint3d}, leverage existing diffusion models along with plug-and-play image prompt module IP-Adapter~\cite{ipadapter}. We propose two potential solutions for incorporating image prompts into MVPaint. One approach is to train a Low-Rank Adaptation (LoRA)~\cite{lora} model akin to CLAY~\cite{clay}, enabling image prompt functionality. Alternatively, we could directly replace the foundational model of T2MV with an image-to-multiview (I2MV) model, such as ImageDream~\cite{imagedream}.


\end{document}